\documentclass[journal]{IEEEtran}

\ifCLASSINFOpdf
\else
\fi

\usepackage{times}
\usepackage{epsfig}
\usepackage{graphicx}
\usepackage{amsmath}
\usepackage{amssymb}
\usepackage{multirow}
\usepackage{url}
\usepackage{subfigure}
\usepackage{subfloat}

\DeclareMathOperator*{\argmin}{arg\,min}

\newcommand{\rulev}{\unskip\ \vrule\ }
\newcommand{\ruleh}{\unskip\ \hrule\ }

\usepackage[pagebackref=true,breaklinks=true,letterpaper=true,colorlinks,bookmarks=false]{hyperref}

\begin{document}
\title{A Unified Framework for Generalizable Style Transfer: Style and Content Separation}

\author{Yexun Zhang,~\IEEEmembership{Student Member,~IEEE,}
        Ya Zhang,~\IEEEmembership{Member,~IEEE,}
        and~Wenbin Cai,~\IEEEmembership{Member,~IEEE}% <-this % stops a space
\thanks{Yexun Zhang and Ya Zhang are with the Cooperative Medianet Innovation Center, Shanghai Jiao Tong University, Shanghai,
China, 200240.\protect\\
E-mail: zhyxun@sjtu.edu.cn, ya$\_$zhang@sjtu.edu.cn}% <-this % stops a space
\thanks{Wenbin Cai is with Microsoft, Beijing, China, 10010.\protect\\
E-mail: wenbca@microsoft.com}% <-this % stops a space
%\thanks{Manuscript received April 25, 2018; revised July 26, 2018.}
}


%\markboth{Journal of \LaTeX\ Class Files,~Vol.~14, No.~8, May~2018}%

\maketitle

\begin{abstract}
Image style transfer has drawn broad attention in recent years. However, most existing methods aim to explicitly model the transformation between different styles, and the learned model is thus not generalizable to new styles. We here propose a unified style transfer framework for both character typeface transfer and neural style transfer tasks leveraging style and content separation. A key merit of such framework is its generalizability to new styles and contents. The overall framework consists of style encoder, content encoder, mixer and decoder. The style encoder and content encoder are used to extract the style and content representations from the corresponding reference images. The mixer integrates the above two representations and feeds it into the decoder to generate images with the target style and content. During training, the encoder networks learn to extract styles and contents from limited size of style/content reference images. This learning framework allows simultaneous style transfer among multiple styles and can be deemed as a special `multi-task' learning scenario. The encoders are expected to capture the underlying features for different styles and contents which is generalizable to new styles and contents. Under this framework, we design two individual networks for character typeface transfer and neural style transfer, respectively. For character typeface transfer, to separate the style features and content features, we leverage the conditional dependence of styles and contents given an image. For neural style transfer, we leverage the statistical information of feature maps in certain layers to represent style. Extensive experimental results have demonstrated the effectiveness and robustness of the proposed methods.
\end{abstract}


\IEEEpeerreviewmaketitle

\section{Introduction}

\IEEEPARstart{I}{n} recent years, style transfer, as an interesting application of deep neural networks (DNNs), has attracted increasing attention among the research community. Based on the type of styles, style transfer may be partitioned into two types of applications, character typeface transfer which transfers a character from a font to another, and neural style transfer which aims to transform a neural image into a given art style. Character typeface transfer usually involves changes in high-frequency features such as the object shape and outline, which makes character typeface transfer a more difficult task than neural style transfer. Moreover, the characters are associated with clear semantic meaning and incorrect transformation may lead to non-sense characters. Different from character typeface transfer, neural style transfer is mostly about the transfer of texture, where the source and target images usually share high-frequency features such as object shape and outline, namely the contents are kept visually unchanged.

Earliest studies about character typeface transfer are usually based on manually extracted features such as radicals and strokes~\cite{lian2016automatic,xu2009automatic,zhang2018drawing,zhou2011easy}.
%For example, most Chinese characters are compound characters, with multiple radicals organized into a specific layout structure~\cite{zhou2011easy}.
Recently, some studies try to automatically learn the transformation based on DNNs, and model character typeface transfer as an image-to-image translation problem. Typically, dedicated models are built for each source and target style pair~\cite{rewrite,lyu}, making the models hardly generalizable to new styles, i.e., additional models have to be trained for new styles. To achieve typeface transfer without retraining, a multi-content generative adversarial networks (GAN) which transfers the font of English characters given a few characters in target styles is proposed~\cite{azadi2017multi}.

Earliest studies for neural style transfer usually adopt an iterative optimization mechanism to generate images with target style and content from noise images~\cite{Gatys}. Due to its time inefficiency, a feed-forward generator network is proposed for this purpose~\cite{johnson,ulyanov}. A set of losses are proposed for the transfer network, such as pixel-wise loss~\cite{isola}, perceptual loss~\cite{johnson,zhang2017multi}, and histogram loss~\cite{wilmot}. Recently, variations of GANs~\cite{liu2016,Zhu2017} are introduced by adding a discriminator to the transfer network which incorporates adversarial loss with transfer loss to generate better images. However, these studies aim to explicitly learn the transformation from a content image to the image with a specific style, and the learned model is thus not generalizable to new styles. So far, there is still limited work for arbitrary neural style transfer~\cite{chen2016fast,Huang_2017_ICCV,li2017universal}.

\begin{table*}[!t]
    \centering
    \caption{Comparison of {\em EMD} with existing methods.}
    \begin{tabular}{c|c|c|c|c}
    \hline
        Methods &  Data format & Generalizable to new styles? & Requirements for new style & What the model learned? \\
    \hline Pix2pix~\cite{isola} & paired & \multirow{8}{3.2cm}{The learned model can only transfer images to styles which appeared in the training set. For new styles, the model has to be retrained.}
    & \multirow{6}{3cm}{Retrain on a lot of training images for a source style and a target style.}  & \multirow{6}{4.2cm}{The translation from a certain source style to a specific target style.}\\
    \cline{1-2} CoGAN~\cite{liu2016} &  unpaired &  & & \\
    \cline{1-2} CycleGAN~\cite{Zhu2017} &  unpaired  & &  \\
    \cline{1-2} Rewrite~\cite{rewrite} &  paired  &  & &\\
    \cline{1-2} Zi-to-zi~\cite{zitozi} & paired  &  & &\\
    \cline{1-2} AEGN~\cite{lyu} & paired  &   & & \\
    \cline{1-2} \cline{4-5} Perceptual~\cite{johnson} & unpaired  &  &  \multirow{3}{3cm}{Retrain on many input content images and one style image.} &
    \multirow{3}{4.2cm}{Transformation among specific styles.}\\
    \cline{1-2} TextureNet~\cite{ulyanov2017improved} & unpaired & & & \\
    \cline{1-2} StyleBank~\cite{chen2017} & unpaired  &  & & \\
    \hline  Patch-based~\cite{chen2016fast} & unpaired  & \multirow{4}{3.2cm}{The learned model can be generalized to new styles.}  & \multirow{4}{3cm}{One or a small set of style/content reference images.} & The swap of style/content feature maps. \\
    \cline{1-2} \cline{5-5} AdaIn~\cite{Huang_2017_ICCV} & unpaired & & & The transferring of feature statistics. \\
    \cline{1-2} \cline{5-5} Universal~\cite{li2017universal} & unpaired & & & It is based on whitening and coloring transformations.\\
    \cline{1-2} \cline{5-5} EMD & triplet/unpaired &  &  & The feature representation of style/content. \\
    \hline
    \end{tabular}%
    \vspace{-10pt}
  \label{tab:comp_methods}%
\end{table*}%

In this paper, based on our previous work~\cite{zhang2018separating}, we propose a unified style transfer framework for both character typeface transfer and neural style transfer, which enables the transfer models generalizable well to new styles or contents. Different from existing style transfer methods, where an individual transfer network is built for each pair of style transfer, the proposed framework represents each style or content with a small set of reference images and attempts to learn separate representations for styles and contents. Then, to generate an image of a given style-content combination is simply to mix the corresponding two representations. This learning framework allows simultaneous style transfer among multiple styles and can be deemed as a special `multi-task' learning scenario. Through separated style and content representations, the framework is able to generate images of all style-content combination given the corresponding reference sets, and is therefore expected to generalize well to new styles and contents.
To our best knowledge, the study most resembles to ours is the bilinear model proposed by Tenenbaum and Freeman~\cite{Tenenbaum}, which obtained independent style and content representations through matrix decomposition. However, to obtain accurate decomposition of new styles and contents, the bilinear model requires an exhaustive enumeration of examples which may not be readily available for some styles/contents.

\begin{figure}[!tbp]
\setlength{\abovecaptionskip}{-2pt}
\centering
\hspace{-10pt}
\includegraphics[height=1.2in,width=3.5in]{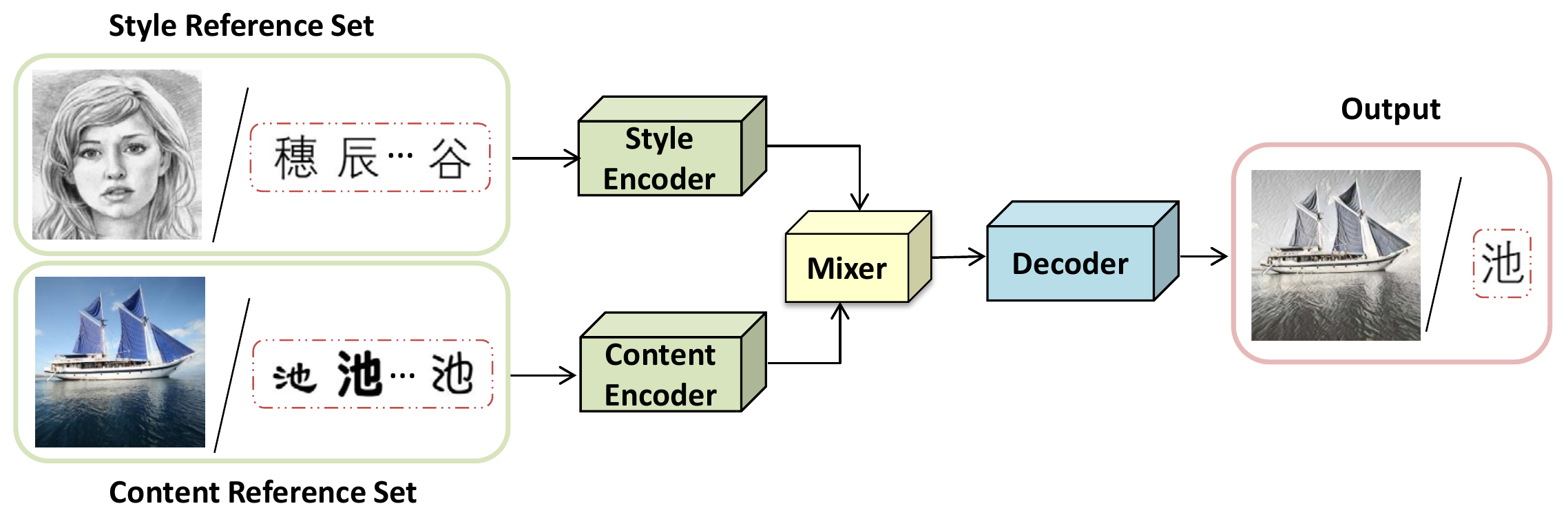}
\caption{The framework of the proposed {\em EMD} model.}
\label{fig:framework1}
\vspace{-15pt}
\end{figure}

As shown in Figure~\ref{fig:framework1}, the proposed style transfer framework, denoted as {\em EMD} thereafter, consists of a style encoder, a content encoder, a mixer, and a decoder. Given one or a set of reference images, the style encoder and content encoder are used to extract the style and content factors from the style reference images and content reference images, respectively. The mixer then combines the corresponding style and content representations. Finally, the decoder generates the target images based on the combined representations. Under this framework, we design two individual networks for character typeface transfer and neural style transfer, respectively. For character typeface transfer, to separate the style features and content features, we leverage the conditional dependence of styles and contents given an image and employ a bilinear model to mix the two factors. For neural style transfer, we leverage the prior knowledge that the statistical information of feature maps in certain layers can represent style information and mix the two factors through statistic matching.

During training, each training example for the proposed network is provided as a style-content pair $<$$\mathcal{R}_{S_i}$, $\mathcal{R}_{C_j}$$>$, where  $\mathcal{R}_{S_i}$ and $\mathcal{R}_{C_j}$ are the style and content reference sets respectively, each consisting of $r$ images of the corresponding style $S_i$ and content $C_j$. For character typeface transfer, the entire network is trained end-to-end with a weighted $L1$ loss measuring the difference between the generated images and the target images. For neural style transfer, due to the absence of target images for supervision, we calculate the content loss and style loss respectively by comparing the feature maps of generated images with those of style/content reference image. Therefore, neural style transfer is unsupervised.
Moreover, due to the difficulty of obtaining images of the same content or style, only one style and content reference image is used as input (namely $r$=1). Extensive experimental results have demonstrated the effectiveness and robustness of our method for style transfer.

The main contributions of our study are summarized as follows.
\begin{itemize}
\item We propose a unified style transfer framework for both character typeface transfer and neural style transfer, which learns separate style and content representations.
\item The framework enables the transfer models generalizable to any unseen style/content given a few reference images.
\item Under this framework, we design two individual networks for character typeface transfer and neural style transfer, respectively, which have shown promising results in experimental validation.
\item This learning framework allows simultaneous style transfer among multiple styles and can be deemed as a special `multi-task' learning scenario.
\end{itemize}

\begin{figure*}[!tbp]
\setlength{\abovecaptionskip}{-2pt}
\centering
\includegraphics[height=2.6in,width=7.2in]{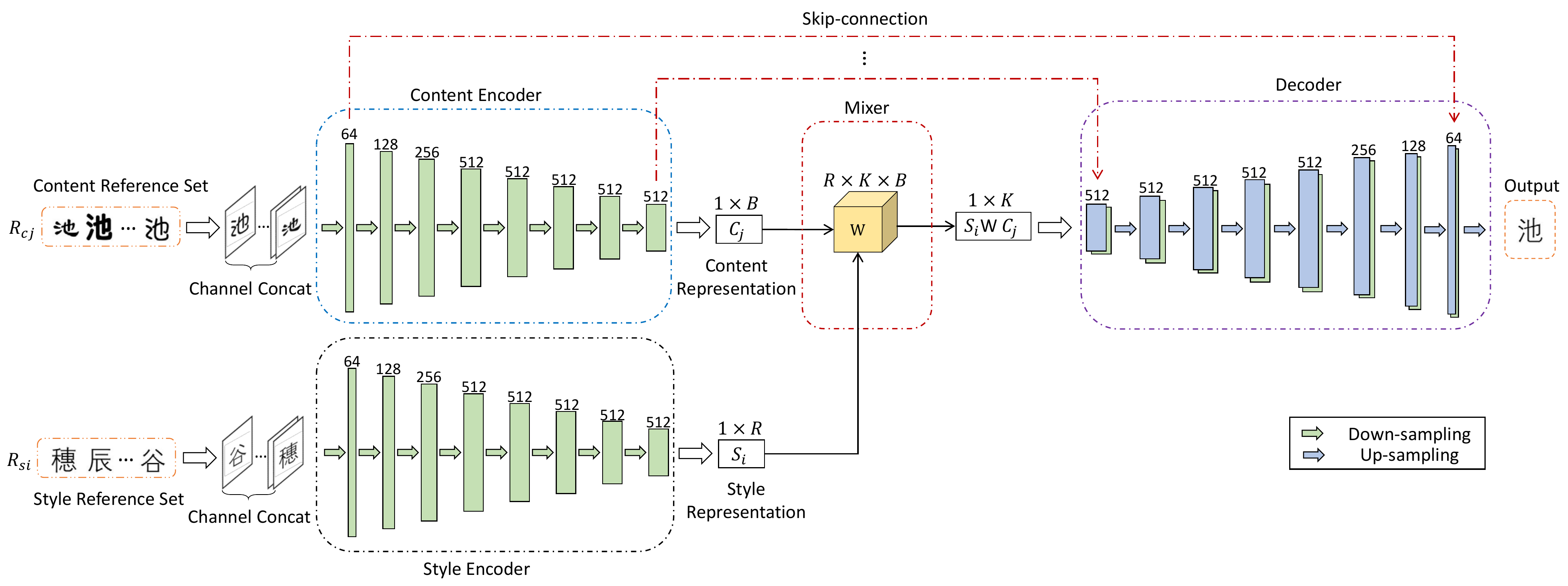}
\caption{The detailed architecture of the proposed generalized {\em EMD} model for character typeface transfer.}
\label{fig:framework2}
\vspace{-15pt}
\end{figure*}

\section{Related Work}

\noindent {\bf Neural Style Transfer.} DeepDream~\cite{mordvintsev} may be considered as the first attempt to generate artistic work using Convolution Neural Networks (CNNs). Gatys et. al later successfully applied CNNs to neural style transfer~\cite{Gatys}. The target images are generated by iteratively optimizing a noise image through a pre-trained network, which is time-consuming.
%Therefore, many studies have been done
To directly learn a feed-forward generator network for neural style transfer, the perceptual loss is proposed~\cite{johnson}. Ulyanov et. al proposed a texture network for both texture synthesis and style transfer~\cite{ulyanov}. Further, Chen et. al proposed the stylebank to represent each style by a convolution filter, which can simultaneously learn numerous styles~\cite{chen2017}. For arbitrary neural style transfer, \cite{chen2016fast} proposed a patch-based method to replace each content feature patch with the nearest style feature. Further, \cite{Huang_2017_ICCV} proposed a faster method based on adaptive instance normalization which performed style transfer in the feature space by transferring feature statistics. Li et. al~\cite{li2017universal} proposed a universal style transfer model which is based on the whitening and coloring transforms but this model is not effective at producing sharp details
and fine strokes.

\noindent {\bf Image-to-Image Translation.} Image-to-image translation is to learn the mapping from the input image to output image, such as from edges to real objects. Pix2pix~\cite{isola} used a conditional GAN based network which requires paired data for training. However, paired data are hard to collect in many applications. Therefore, methods requiring non-paired data are explored. Liu and Tuzel proposed the coupled GAN (CoGAN)~\cite{liu2016} to learn a joint distribution of two domains through weight sharing. Later, Liu~\cite{liu2017} extended the CoGAN to unsupervised image-to-image translation. Some other studies~\cite{Bousmalis,Shrivastava,Taigmand} encourage the input and output to share certain content even though they may differ in style by enforcing the output to be close to the input in a predefined metric space such as class label space. Recently, Zhu et. al proposed the cycle-consistent adversarial network (CycleGAN)~\cite{Zhu2017} which performs well for many vision and graphics tasks.

\noindent {\bf Character Typeface Transfer.} Most existing studies model character typeface transfer as an image translation task. The ``Rewrite" project uses a simple top-down CNNs structure and transfers a typographic font to another stylized typographic font~\cite{rewrite}. As the improvement version, the ``zi-to-zi" project can transfer multiple styles by assigning each style an one-hot category label and training the network in a supervised way~\cite{zitozi}. The recent work ``From A to Z" also adopts a supervised method and assigns each character an one-hot label~\cite{upchurch}. Lyu et. al proposed an auto-encoder guided GAN network (AEGN) which can synthesize calligraphy images with specified style from standard Chinese font images~\cite{lyu}.~\cite{azadi2017multi} proposed a multi-content GAN which could achieve typeface transfer on English characters with a few examples of target style.

However, existing work usually studies character typeface transfer and neural style transfer individually, while the proposed {\em EMD} provides a unified framework which is applicable to both tasks. In addition, most of the methods reviewed above can only transfer styles in the training set and the network must be retrained for new styles. In contrast, the proposed {\em EMD} framework can generate images with new styles/contents given only a few of reference images. We present a comparison of the methods in Table~\ref{tab:comp_methods}.

\section{Generalized Style Transfer Framework}

The generalized style transfer framework {\em EMD} is an encoder-decoder network which consists of four subnets: $\textsl{Style Encoder}$, $\textsl{Content Encoder}$, $\textsl{Mixer}$ and $\textsl{Decoder}$, as shown in Figure~\ref{fig:framework1}. First, the $\textsl{Style/Content Encoder}$ extracts style/content representations given style/content reference images. Next, the $\textsl{Mixer}$ integrates the style feature and content feature, and the combined feature is then fed into the $\textsl{Decoder}$. Finally, the $\textsl{Decoder}$ generates the image with the target style and content.

The input of the $\textsl{Style Encoder}$ and $\textsl{Content Encoder}$ are style reference set $\mathcal{R}_{S_i}$ and content reference set $\mathcal{R}_{C_j}$, respectively. $\mathcal{R}_{S_i}$ consists of $r$ reference images with the same style $S_i$ but different contents ${C_{j_1},C_{j_2},\ldots,C_{j_r}}$
\begin{equation}
\mathcal{R}_{S_i} = \{I_{ij_1}, I_{ij_2}, \ldots, I_{ij_r} \},
\end{equation}
where $I_{ij}$ represents the image with style $S_i$ and content $C_j$. For example, in character typeface transfer tasks, $\mathcal{R}_{S_i}$ contains $r$ images with the same font $S_i$ such as serif, sanserif, and blackletter, but different characters.
Similarly, $\mathcal{R}_{C_j}$ is for content $C_j$ $(j=1,2,\ldots,m)$ which consists of $r$ reference images of the same character $C_j$ but in different styles ${S_{i_1},S_{i_2},\ldots,S_{i_r}}$
\begin{equation}
\mathcal{R}_{C_j} = \{I_{i_1j}, I_{i_2j}, \ldots, I_{i_rj} \}.
\end{equation}
The whole framework is trained end-to-end by trying to finish a series of tasks: generate images with target style and content given the style and content reference images. By such a way, we expect the framework to summarize from these similar tasks and learn to extract style and content representations, and then transfer this ability to new styles and contents.

It is worth noting that the proposed {\em EMD} learning framework is quite flexible and the $\textsl{Style Encoder}$, $\textsl{Content Encoder}$, $\textsl{Mixer}$, and $\textsl{Decoder}$ can be tailored based on specific tasks. %Besides, the mixing method adopted by the $\textsl{Mixer}$ can also be different, indicating that our framework is flexible and can be applied to different tasks.
In the rest of the section, under this framework, we demonstrate with two individual networks for character typeface transfer and neural style transfer, respectively.

\section{Character Typeface Transfer}
The detailed network architecture employed for character typeface transfer is shown in Figure~\ref{fig:framework2}.

\subsection{Encoder Network}
The two encoder networks used for character typeface transfer have the same architecture and consist of a series of Convolution-BatchNorm-LeakyReLU down-sampling blocks which yield 1$\times$1 feature representations of the input style/content reference images. The first convolution layer is with 5$\times$5 kernel and stride 1 and the rest are with 3$\times$3 kernel and stride 2. All ReLUs are leaky, with slope 0.2. The $r$ input reference images are concatenated in the channel dimension to feed into the encoders. This allows the encoders to capture the common characteristics among images of the same style/content.

\subsection{Mixer Network}
Given the style representations and content representations obtained by the $\textsl{Style Encoder}$ and $\textsl{Content Encoder}$, we employ a bilinear model as the $\textsl{Mixer}$ to combine the two factors. The bilinear models are two-factor models with the mathematical property of separability: their outputs are linear in either factor when the other is held constant. It has been demonstrated that the influences of the two factors can be efficiently separated and combined in a flexible representation that can be naturally generalized to unfamiliar factor classes such as new styles~\cite{Tenenbaum}. Furthermore, the bilinear model has also been successfully used in zero-shot learning as a compatibility function to associate visual representation and auxiliary class text description~\cite{changpinyo,frome,xian}. The learned compatibility function can be seen as the shared knowledge and transferred to new classes. Here, we take the bilinear model to integrate styles and contents together which is formulated as
\begin{equation}
F_{ij} = S_i\textbf{W}C_j,
\end{equation}
where $\textbf{W}$ is a tensor with size $R\times K\times B$, $S_i$ is the $R$-dimensional style feature and $C_j$ is the $B$-dimensional content feature. $F_{ij}$ can be seen as the $K$-dimensional feature vector of image $I_{ij}$ which is further taken as the input of the $\textsl{Decoder}$ to generate the image with style $S_i$ and content $C_j$.

\subsection{Decoder Network}

The image generator is a typical decoder network which is symmetrical to the encoder and maps the combined feature representation to output images with target style and content. The $\textsl{Decoder}$ roughly follows the architectural guidelines set forth by Radford et. al~\cite{Radford} and consists of a series of Deconvolution-BatchNorm-ReLU up-sampling blocks except that the last layer is the deconvolution layer. Other than the last layer which uses 5$\times$5 kernels and stride 1, all deconvolution layers use 3$\times$3 kernels and stride 2. The outputs are finally transformed into [0,1] by the sigmoid function.

In addition, because the stride convolution in $\textsl{Style Encoder}$ and $\textsl{Content Encoder}$ is detrimental to the extraction of spatial information, we adopt the skip-connection which has been commonly used in semantic segmentation tasks~\cite{jegou,long,ronneberger} to refine the segmentation using spatial information from different resolutions. Although the content inputs and outputs differ in appearance, they share the same structure. Hence, we concatenate the input feature map of each up-sampling block with the corresponding output of the symmetrical down-sampling block in $\textsl{Content Encoder}$ to allow the $\textsl{Decoder}$ to learn back the relevant structure information lost during the down-sampling process.

\begin{figure*}[!t]
\setlength{\abovecaptionskip}{-2pt}
\centering
\includegraphics[height=2.1in,width=7.2in]{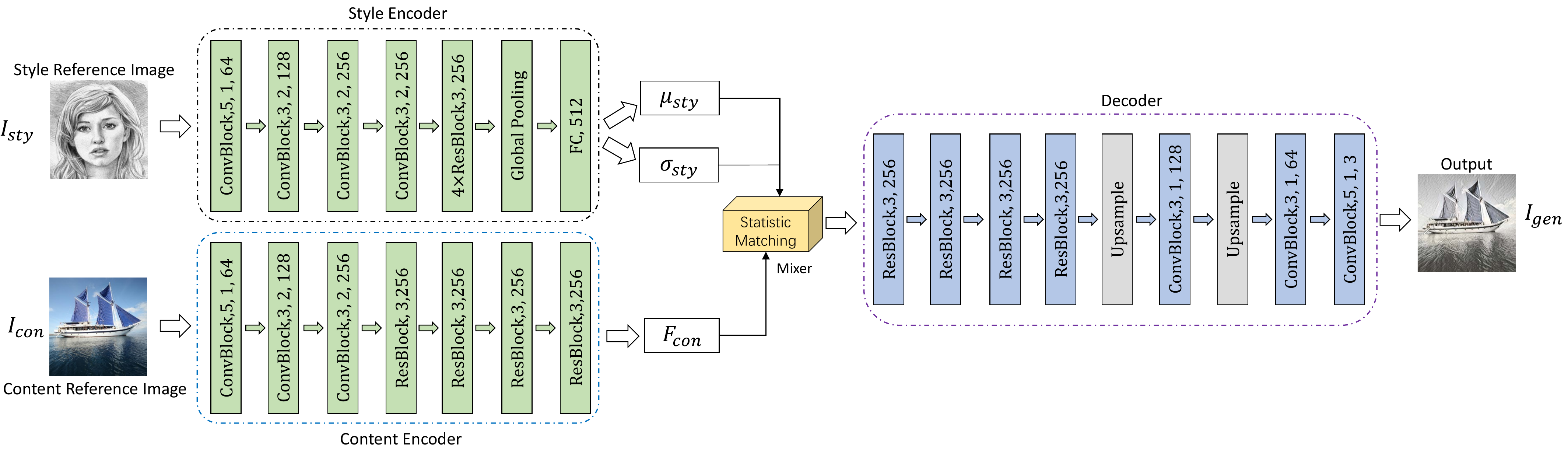}
\caption{The detailed architecture of the proposed generalized {\em EMD} model for neural style transfer.}
\label{fig:framework3}
\vspace{-15pt}
\end{figure*}

\subsection{Loss Function}
For character typeface transfer tasks, it is possible to obtain a reasonable set of the target images. Therefore, we leverage the target images to train the network. Given a training set $\mathcal{D}_t$, the training objective is defined as
\begin{alignat}{3}
\theta = \argmin_{\theta} \sum_{I_{ij} \in \mathcal{D}_t} \mathcal{L}(\hat{I}_{ij},I_{ij}|\mathcal{R}_{S_i},\mathcal{R}_{C_j}; \theta),
\label{eq:model}
\end{alignat}
where $\theta$ represents the model parameters, $\hat{I}_{ij}$ is the generated image and $\mathcal{L}(\hat{I}_{ij},I_{ij}|\mathcal{R}_{S_i},\mathcal{R}_{C_j}; \theta)$ is the generation loss which is formulated as
\begin{equation}
\mathcal{L}(\hat{I}_{ij},I_{ij}|\mathcal{R}_{S_i},\mathcal{R}_{C_j}; \theta) = W^{ij}_{st}\times W^{ij}_d \times ||\hat{I}_{ij} - I_{ij}||.
\end{equation}
The pixel-wise L1 loss is employed as the generation loss for character typeface transfer problem rather than L2 loss because L1 loss tends to yield sharper and cleaner images~\cite{isola,lyu}.

%$W^{ij}_{st}$ and $W^{ij}_b$ are two weights for target image $I_{ij}$ which are added to alleviate the imbalance in the target set induced by the random sampling.
In each learning iteration, the size, thickness, and darkness of the characters in the target set may vary significantly. Due to the way the loss is defined, the model tends to optimize for characters with more pixels, i.e., big and thick characters. Moreover, models trained using L1 loss tend to pay more attention to darker characters and perform poorly on lighter characters. To alleviate the above imbalance, we add two weights to the generation loss: $W^{ij}_{st}$ about the size and thickness of characters, and $W^{ij}_d$ about the darkness of characters.

As for $W^{ij}_{st}$, we first calculate the number of black pixels, i.e., pixels whose values are less than 0.5 after normalized into [0,1]. Then $W^{ij}_{st}$ is defined as the reciprocal of the number of black pixels in each target image
\begin{equation}
W^{ij}_{st} = 1/N_b^{ij},
\end{equation}
where $N_b^{ij}$ is the number of black pixels of target image $I_{ij}$.

As for $W^{ij}_d$, we calculate the mean value of black pixels for each target image and set a softmax weight
\begin{equation}
W^{ij}_d = \frac{exp(\textrm{mean}_{ij})}{\sum_{I_{ij}\in \mathcal{D}_t}exp(\textrm{mean}_{ij})},
\end{equation}
where $\textrm{mean}_{ij}$ is the mean value of the black pixels of the target image $I_{ij}$.

\section{Neural Style Transfer}
We further apply the {\em EMD} framework to neural style transfer. Due to the difficulty of finding neural images with the same style or content, the input to the $\textsl{Style Encoder}$ and $\textsl{Content Encoder}$ is one image. For simplicity, we denote the style image $I_{sty}$ and the content image $I_{con}$.

Many existing neural style transfer methods employ the Gram matrix to represent styles \cite{Gatys,johnson} and style transfer is achieved by matching the Gram matrix of generated images with that of style images. It has been theoretically proved that if we consider the activation at each position of feature maps as individual samples, then matching Gram matrix can be reformulated as minimizing the Maximum Mean Discrepancy (MMD)~\cite{Li2017Demystifying}. Therefore, neural style transfer can be seen as distribution alignment from the content image to the style image~\cite{Li2017Demystifying}.

Based on above foundation, the Conditional Instance Normalization (CIN) method proposes to learn a set of affine parameters ($\gamma_s$ and $\beta_s$) for each style and transfers style with an affine transformation \cite{dumoulin2016learned}
\begin{equation}
    \hat{F} = \frac{F_{con} - \mu(F_{con})}{\sigma(F_{con})} \gamma_s + \beta_s,
\end{equation}
where $F_{con}$ are the feature maps of the content reference image, $\mu(F_{con})$ and $\sigma(F_{con})$ are the mean and standard deviation of $F_{con}$ across the spatial axis. Despite of its promising performance, this method is restricted to styles in the training set. To solve this problem, \cite{Huang_2017_ICCV} designed an Adaptive Instance Normalization (AdaIN) layer where the affine parameters are directly calculated from the style feature maps of a certain layer in pre-trained VGG-19, namely $\gamma_s$=$\sigma(F_{sty})$ and $\beta_s$=$\mu(F_{sty})$. But this is not as accurate as CIN because the calculated affine parameters are indeed estimation of the real statistics. Borrowing ideas from the above two studies, our method learns the affine parameters from the style image by the $\textsl{Style Encoder}$, which is both flexible and accurate.

\subsection{Network Architecture}

For neural style transfer, the $\textsl{Style Encoder}$ consists of a stack of Convolution Blocks and Residual Blocks, a Global Pooling layer and a Fully-Connected layer. Each Convolution Block $<$$ConvBlock$,$k$,$s$,$c$$>$ is composed of a convolution layer with kernel size $k$, stride $s$ and filter number $c$ and a LeakyReLU layer with slope 0.2. Each Residual block $<$$ResBlock$,$k$,$c$$>$ consists of two convolution blocks $<$$ConvBlock$,$k$,$1$,$c$$>$. Then the Global Pooling layer (here we use Global Average Pooling) produces a feature map of size $1\times1$. The final Fully-Connected layer $<$$FC$,$c$$>$ is used to generate the $c$-dimensional statistic vectors (mean and standard deviation). For $\textsl{Content Encoder}$, we use three Convolution Blocks followed by four Residual Blocks. The detailed network architecture is displayed in Figure~\ref{fig:framework3}.

Through the $\textsl{Content Encoder}$, we obtain the feature maps $F_{con}$ of the content reference image $I_{con}$. In addition, the distribution statistics of the style reference image $I_{sty}$ are learned by the $\textsl{Style Encoder}$ and we denote the mean by $\mu_{sty}$ and the standard deviation by $\sigma_{sty}$. Then based on the foundation that neural style transfer can be seen as a distribution alignment process from the content image to the style image, we mix these two factors by statistic matching between style and content images
\begin{equation}
    \hat{F}^c = \frac{F_{con}^c-\mu(F_{con}^c)}{\sigma(F_{con}^c)} \sigma_{sty}^c + \mu_{sty}^c,
\end{equation}
where $\hat{F}^c$ is the statistic aligned feature map for the $c$-th channel. $\mu(F_{con}^c)$ and $\sigma(F_{con}^c)$ are the mean and standard deviation computed across all positions of feature map $F_{con}^c$
\begin{equation}
    \mu(F_{con}^c) = \frac{1}{HW} \sum_{h=1}^H \sum_{w=1}^W F_{con}^{hwc},
\end{equation}
\begin{equation}
    \sigma(F_{con}^c) = \left[\frac{1}{HW} \sum_{h=1}^H\sum_{w=1}^W (F_{con}^{hwc} - \mu(F_{con}^c))^2 \right]^\frac{1}{2},
\end{equation}
where we suppose the size of $F_{con}$ is $H\times W \times C$.

The $\textsl{Decoder}$ takes the feature maps $\hat{F}$ as the input and generates the image $I_{gen}$ with target style and content. The architecture of the $\textsl{Decoder}$ mostly mirrors the layers of $\textsl{Content Encoder}$ except that the stride-2 convolution is replaced by stride-1 convolution and each convolution layer is followed by a ReLU rectifier except the last layer. Besides, we upsample the feature maps by nearest neighbor method in up-sample layers to reduce checkerboard effects as done in~\cite{Huang_2017_ICCV}.

\subsection{Loss Function}
Similar to \cite{ulyanov}, we use a pretrained VGG-19 model to calculate the loss function
\begin{equation}
   \mathcal{L}(I_{gen}|I_{sty},I_{con}) = \lambda_c \mathcal{L}_c + \lambda_s \mathcal{L}_s + \lambda_{tv} \mathcal{L}_{tv},
\end{equation}
which is a weighted combination of the content loss $\mathcal{L}_c$, the style loss $\mathcal{L}_s$ and the total variation regularizer $\mathcal{L}_{tv}$.

The content loss $\mathcal{L}_c$ is the squared and normalized Euclidean distance between the feature maps of generated images and content reference images. Suppose the content loss is calculated for the $l$-th layer and the feature maps are of size $H_l\times W_l \times C_l$, then the content loss can be formulated as
\begin{equation}
    \mathcal{L}_c = \frac{1}{H_lW_lC_l}\parallel F^l_{gen} - F^l_{con} \parallel _2^2,
\end{equation}
where $F^l_{gen}$ and $F^l_{con}$ are feature maps in the $l$-th layer for the generated image $I_{gen}$ and the content reference image $I_{con}$.

The style loss $\mathcal{L}_s$ is constructed by aligning the Batch Normalization (BN) statistics (mean and standard deviation) \cite{Huang_2017_ICCV,Li2017Demystifying} of the feature maps of the generated image $I_{gen}$ and the style reference image $I_{sty}$
\begin{alignat}{3}
\mathcal{L}_s = & \sum_{l} \parallel \mu (F^l_{gen}) - \mu(F^l_{sty})
    \parallel_2^2 \nonumber \\
    & + \parallel \sigma (F^l_{gen}) - \sigma(F^l_{sty}) \parallel_2^2.
\end{alignat}
In addition, following~\cite{johnson,Mahendran2014}, we add the total variation regularizer $\mathcal{L}_{tv}$ to encourage the smooth of generated images.

\section{Experiments}

%In this section, we present experiments on character typeface transfer task and neural style transfer task to demonstrate the effectiveness and robustness of our proposed framework.

\subsection{Character Typeface Transfer}
%We evaluate the proposed network on Chinese Typeface transfer problem. We first introduce the data set we used followed by the implementation details. Finally, we present our experimental results.

\begin{figure}[!t]
\setlength{\abovecaptionskip}{-2pt}
\centering
\hspace{-5pt}
\includegraphics[height=1.7in,width=3.3in]{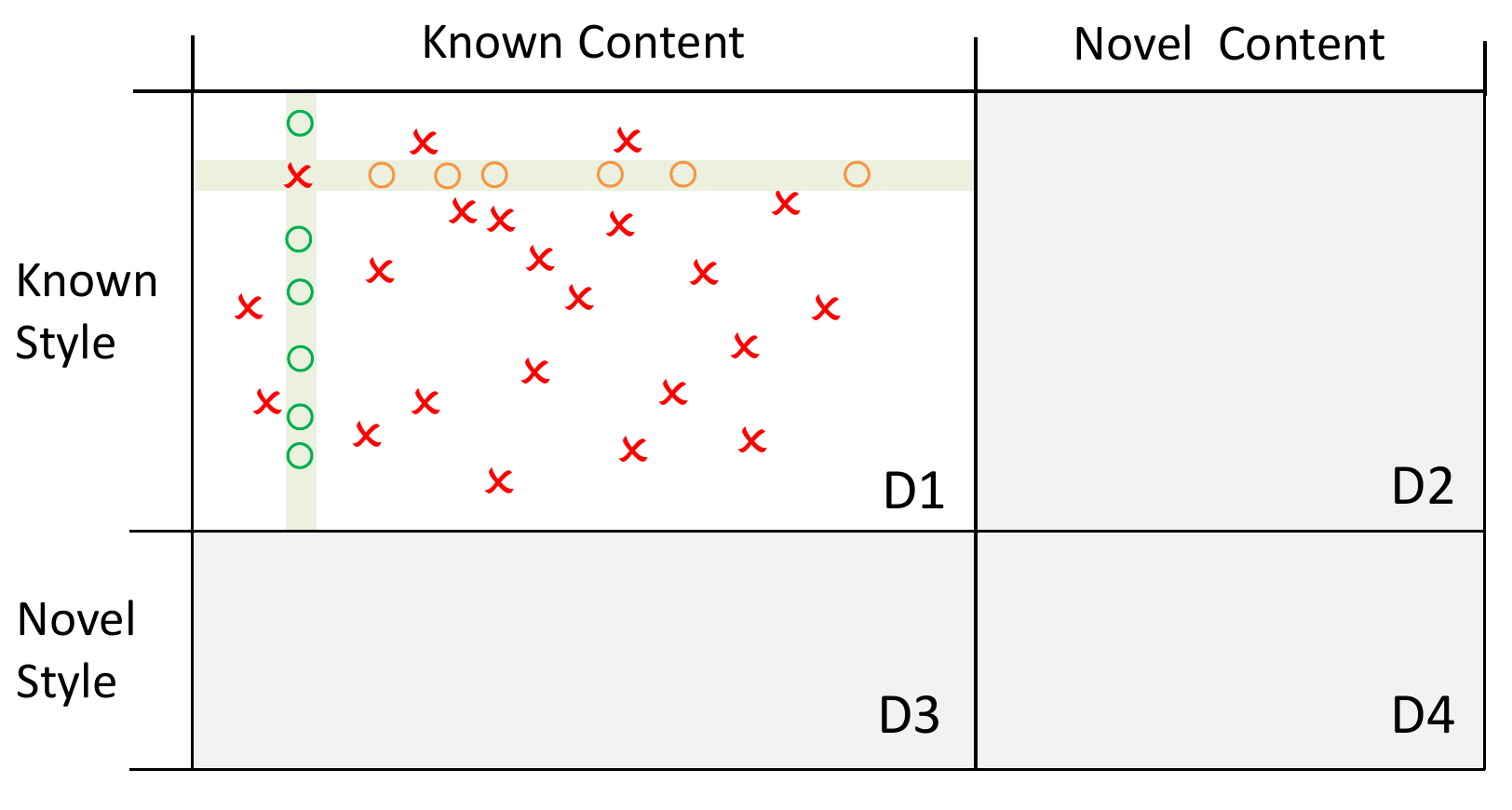}
\caption{The illustration of data set partition, target images selection and reference set construction (best viewed in color).}
\label{fig:train}
\vspace{-5pt}
\end{figure}

\subsubsection{Data Set}
To evaluate the proposed {\em EMD} model with Chinese Typeface transfer tasks, we construct a data set of 832 fonts (styles), each font with 1732 commonly used Chinese characters (contents). All images are in the size of $80\times 80$ pixels. We randomly select 75\% of the styles and contents as known styles and contents (i.e. 624 train styles and 1299 train contents) and leave the rest 25\% as novel styles and contents (i.e. 208 novel styles and 433 novel contents). The entire data set is accordingly partitioned into four subsets as shown in Figure~\ref{fig:train}: $D_1$, images with known styles and contents, $D_2$, images with known styles but novel contents, $D_3$, images with known contents but novel styles, and $D_4$, images with both novel styles and novel contents. The training set is selected from $D_1$, and four test sets are selected from $D_1$, $D_2$, $D_3$, and $D_4$, respectively. The four test sets represent different levels of style transfer challenges.
%***********************************training set size ***************************************************
\begin{figure}[!t]
\centering
\setlength{\abovecaptionskip}{5pt}
\hspace{-6pt}
\subfigure{
\begin{minipage}{0.23\textwidth}{\vspace{-5pt}
\begin{minipage}{0.12\textwidth}TG:\end{minipage}
\begin{minipage}{0.15\textwidth}
\includegraphics[width=1.5in,height=0.21in]{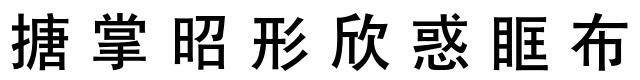}
\end{minipage}\\
\begin{minipage}{0.12\textwidth}O1:\end{minipage}
\begin{minipage}{0.15\textwidth}
\includegraphics[width=1.5in,height=0.21in]{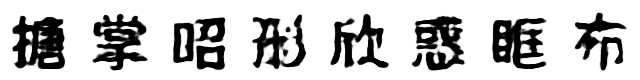}
\end{minipage}\\
\begin{minipage}{0.12\textwidth}O2:\end{minipage}
\begin{minipage}{0.15\textwidth}
\includegraphics[width=1.5in,height=0.21in]{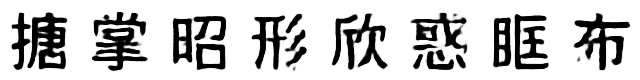}
\end{minipage}\\
\begin{minipage}{0.12\textwidth}O3:\end{minipage}
\begin{minipage}{0.15\textwidth}
\includegraphics[width=1.5in,height=0.21in]{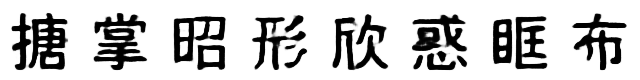}
\end{minipage}\\
\begin{minipage}{0.12\textwidth}O4:\end{minipage}
\begin{minipage}{0.15\textwidth}
\includegraphics[width=1.5in,height=0.21in]{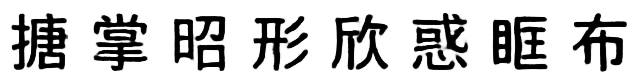}
\end{minipage}\\
\begin{minipage}{0.12\textwidth}O5:\end{minipage}
\begin{minipage}{0.15\textwidth}
\includegraphics[width=1.5in,height=0.21in]{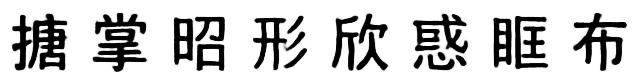}
\end{minipage}
\ruleh
\\
\begin{minipage}{0.12\textwidth}TG:\end{minipage}
\begin{minipage}{0.15\textwidth}
\includegraphics[width=1.5in,height=0.21in]{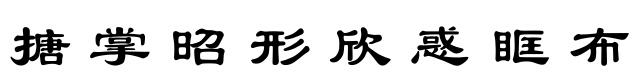}
\end{minipage}\\
\begin{minipage}{0.12\textwidth}O1:\end{minipage}
\begin{minipage}{0.15\textwidth}
\includegraphics[width=1.5in,height=0.21in]{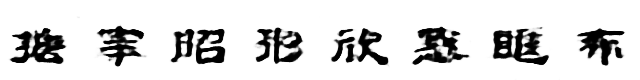}
\end{minipage}\\
\begin{minipage}{0.12\textwidth}O2:\end{minipage}
\begin{minipage}{0.15\textwidth}
\includegraphics[width=1.5in,height=0.21in]{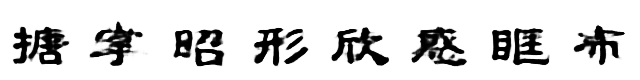}
\end{minipage}\\
\begin{minipage}{0.12\textwidth}O3:\end{minipage}
\begin{minipage}{0.15\textwidth}
\includegraphics[width=1.5in,height=0.21in]{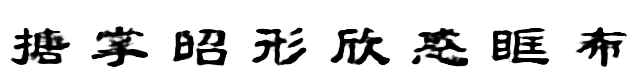}
\end{minipage}\\
\begin{minipage}{0.12\textwidth}O4:\end{minipage}
\begin{minipage}{0.15\textwidth}
\includegraphics[width=1.5in,height=0.21in]{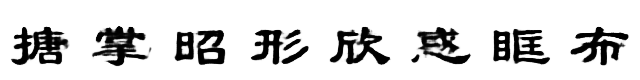}
\end{minipage}\\
\begin{minipage}{0.12\textwidth}O5:\end{minipage}
\begin{minipage}{0.15\textwidth}
\includegraphics[width=1.5in,height=0.21in]{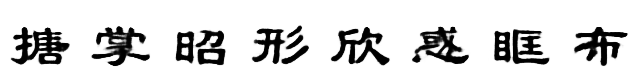}
\end{minipage}
}
\end{minipage}
}
\vspace{-5pt}
\rulev
\hspace{-3pt}
\subfigure{
\begin{minipage}{0.21\textwidth}{\vspace{-6pt}
\includegraphics[width=1.5in,height=0.21in]{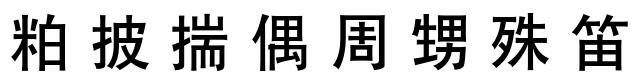}\\
\includegraphics[width=1.5in,height=0.21in]{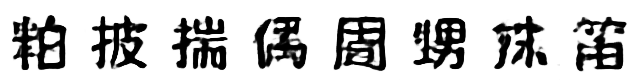}\\
\includegraphics[width=1.5in,height=0.21in]{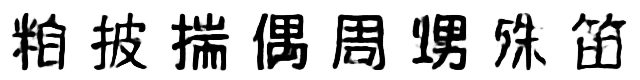}\\
\includegraphics[width=1.5in,height=0.21in]{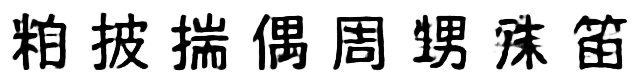}\\
\includegraphics[width=1.5in,height=0.21in]{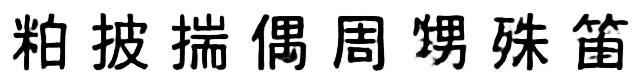}\\
\includegraphics[width=1.5in,height=0.21in]{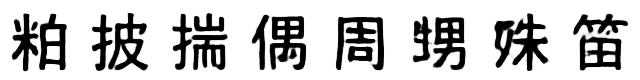}
\ruleh
\\
\includegraphics[width=1.5in,height=0.21in]{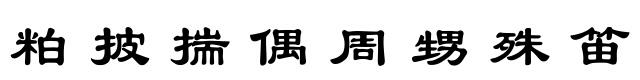}\\
\includegraphics[width=1.5in,height=0.21in]{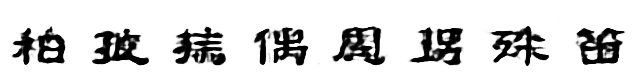}\\
\includegraphics[width=1.5in,height=0.21in]{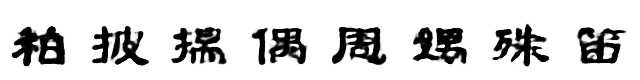}\\
\includegraphics[width=1.5in,height=0.21in]{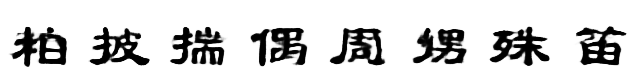}\\
\includegraphics[width=1.5in,height=0.21in]{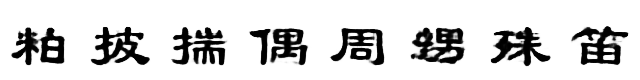}\\
\includegraphics[width=1.5in,height=0.21in]{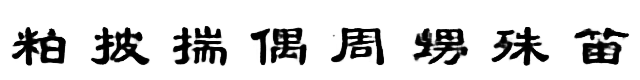}

}
\end{minipage}
}
\caption{Generation results for $D_1$, $D_2$, $D_3$, $D_4$ (from upper left to lower right) with different training set size. TG: Target image, O1: Output for $N_{t}$=20k, O2: Output for $N_{t}$=50k, O3: Output for $N_{t}$=100k, O4: Output for $N_{t}$=300k, O5: Output for $N_{t}$=500k. In all cases, $r$=10.}
\label{fig:train-size}
\vspace{-15pt}
\end{figure}

\subsubsection{Implementation Details}
In our experiment, the output channels of convolution layers in the $\textsl{Style Encoder}$ and $\textsl{Content Encoder}$ are 1, 2, 4, 8, 8, 8, 8, 8 times of $C$ respectively, where $C$=64. And for the $\textsl{Mixer}$, we set $R$=$B$=$K$ in our implementation. The output channels of the first seven deconvolution layers in $\textsl{Decoder}$ are 8, 8, 8, 8, 4, 2, 1 times of $C$ respectively. We set the initial learning rate as 0.0002 and train the model end-to-end with the Adam optimization method until the output is stable.

In each experiment, we first randomly sample $N_{t}$ target images with known content and known styles from $D_1$ as training examples. We then construct the two reference sets for each target image by randomly sampling $r$ images of the corresponding style/content from $D_1$. Figure~\ref{fig:train} provides an illustration of target images selection and reference set construction. Each row represents one style and each column represents a content. The target images are represented by randomly scattered red ``x" marks. The reference images for the target image are selected from corresponding style/content, shown as the orange circles for the style reference images and green circles for content reference images.

\subsubsection{Experimental Results}

%In this subsection, we present the experimental results of Chinese Typeface transfer. First, we analyze the influence of some factors influencing the model performance. Then, we validate the separation of style and content. Finally, we compare the proposed method with some baseline networks to prove the effectiveness of our method.

\noindent {\bf Influence of the Training Set Size }
To evaluate the influence of the training set size on style transfer, we conduct experiments for $N_{t}$=20k, 50k, 100k, 300k and 500k. The generation results for $D_{1}$, $D_{2}$, $D_{3}$ and $D_{4}$ are shown in Figure~\ref{fig:train-size}. As we can see, the larger the training set, the better the performance, which is consistent with our intuition. The generated images with $N_{t}$=300k and 500k are clearly better than images generated with $N_{t}$=20k, 50k and 100k. Besides, the performance of $N_{t}$=300k and $N_{t}$=500k is close which implies that with more training images, the network performance tends to be saturated and $N_{t}$=300k is enough for good results. Therefore, we take $N_{t}$=300k for the rest of experiments.

%***********************************k-shot***************************************************
\begin{figure}[!t]
\centering
\setlength{\abovecaptionskip}{-2pt}
\hspace{-7pt}
\subfigure{
\begin{minipage}{0.23\textwidth}{\vspace{-5pt}
\begin{minipage}{0.12\textwidth}TG:\end{minipage}
\begin{minipage}{0.15\textwidth}
\includegraphics[width=1.5in,height=0.21in]{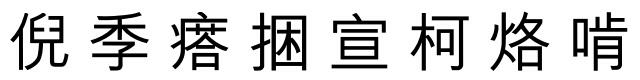}
\end{minipage}\\
% \begin{minipage}{0.12\textwidth}O1:\end{minipage}
% \begin{minipage}{0.15\textwidth}
% \includegraphics[width=1.5in,height=0.21in]{results_300000-comparison_ref_num-5s-1c-test1-24-outputs.png}
% \end{minipage}\\
\begin{minipage}{0.12\textwidth}O1:\end{minipage}
\begin{minipage}{0.15\textwidth}
\includegraphics[width=1.5in,height=0.21in]{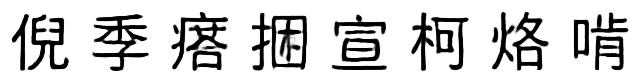}
\end{minipage}\\
\begin{minipage}{0.12\textwidth}O2:\end{minipage}
\begin{minipage}{0.15\textwidth}
\includegraphics[width=1.5in,height=0.21in]{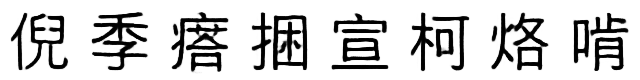}
\end{minipage}\\
\begin{minipage}{0.12\textwidth}O3:\end{minipage}
\begin{minipage}{0.15\textwidth}
\includegraphics[width=1.5in,height=0.21in]{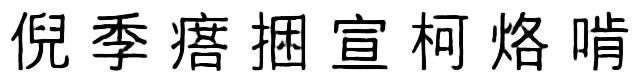}
\end{minipage}
\vspace{3pt}\\
\begin{minipage}{0.12\textwidth}TG:\end{minipage}
\begin{minipage}{0.15\textwidth}
\includegraphics[width=1.5in,height=0.21in]{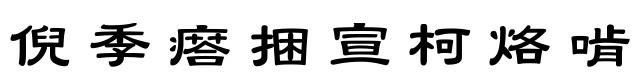}
\end{minipage}\\
\begin{minipage}{0.12\textwidth}O1:\end{minipage}
\begin{minipage}{0.15\textwidth}
\includegraphics[width=1.5in,height=0.21in]{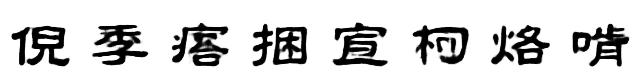}
\end{minipage}\\
\begin{minipage}{0.12\textwidth}O2:\end{minipage}
\begin{minipage}{0.15\textwidth}
\includegraphics[width=1.5in,height=0.21in]{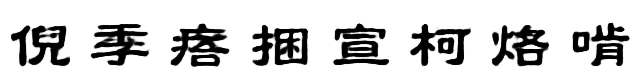}
\end{minipage}\\
\begin{minipage}{0.12\textwidth}O3:\end{minipage}
\begin{minipage}{0.15\textwidth}
\includegraphics[width=1.5in,height=0.21in]{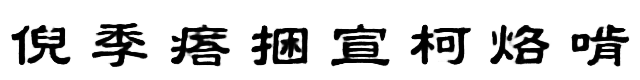}
\end{minipage}
\ruleh
\\
\begin{minipage}{0.12\textwidth}TG:\end{minipage}
\begin{minipage}{0.15\textwidth}
\includegraphics[width=1.5in,height=0.21in]{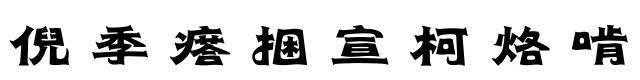}
\end{minipage}\\
\begin{minipage}{0.12\textwidth}O1:\end{minipage}
\begin{minipage}{0.15\textwidth}
\includegraphics[width=1.5in,height=0.21in]{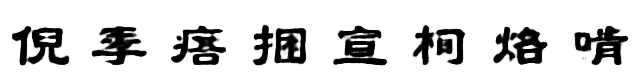}
\end{minipage}\\
\begin{minipage}{0.12\textwidth}O2:\end{minipage}
\begin{minipage}{0.15\textwidth}
\includegraphics[width=1.5in,height=0.21in]{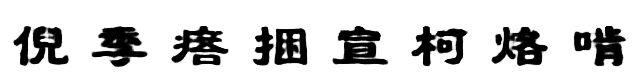}
\end{minipage}\\
\begin{minipage}{0.12\textwidth}O3:\end{minipage}
\begin{minipage}{0.15\textwidth}
\includegraphics[width=1.5in,height=0.21in]{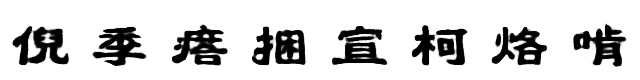}
\end{minipage}
\vspace{3pt}\\
\begin{minipage}{0.12\textwidth}TG:\end{minipage}
\begin{minipage}{0.15\textwidth}
\includegraphics[width=1.5in,height=0.21in]{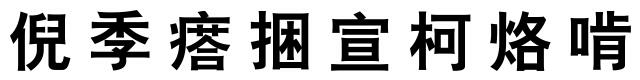}
\end{minipage}\\
\begin{minipage}{0.12\textwidth}O1:\end{minipage}
\begin{minipage}{0.15\textwidth}
\includegraphics[width=1.5in,height=0.21in]{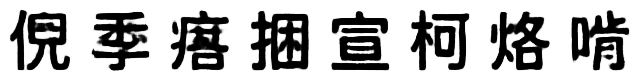}
\end{minipage}\\
\begin{minipage}{0.12\textwidth}O2:\end{minipage}
\begin{minipage}{0.15\textwidth}
\includegraphics[width=1.5in,height=0.21in]{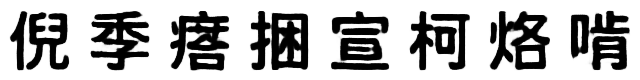}
\end{minipage}\\
\begin{minipage}{0.12\textwidth}O3:\end{minipage}
\begin{minipage}{0.15\textwidth}
\includegraphics[width=1.5in,height=0.21in]{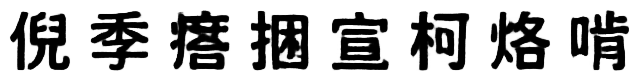}
\end{minipage}
}
\end{minipage}
}
\hspace{1pt}
\rulev
\hspace{-3pt}
\subfigure{
\begin{minipage}{0.21\textwidth}{\vspace{-6pt}
\includegraphics[width=1.5in,height=0.21in]{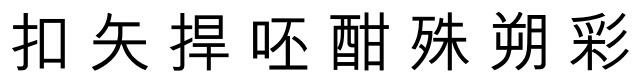}\\
\includegraphics[width=1.5in,height=0.21in]{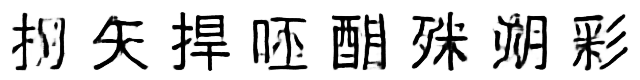}\\
\includegraphics[width=1.5in,height=0.21in]{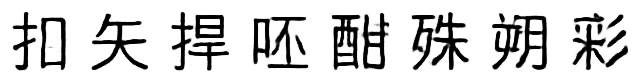}\\
\includegraphics[width=1.5in,height=0.21in]{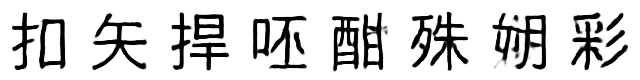}\vspace{3pt}\\
\includegraphics[width=1.5in,height=0.21in]{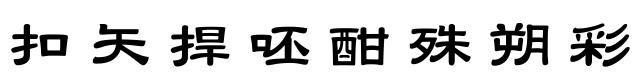}\\
\includegraphics[width=1.5in,height=0.21in]{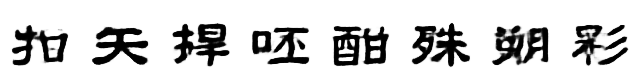}\\
\includegraphics[width=1.5in,height=0.21in]{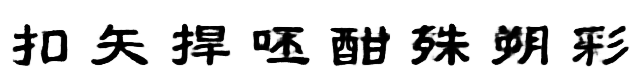}\\
\includegraphics[width=1.5in,height=0.21in]{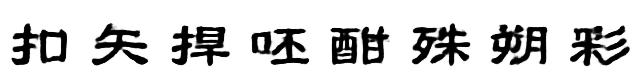}
\ruleh
\vspace{1pt}
\\
\includegraphics[width=1.5in,height=0.21in]{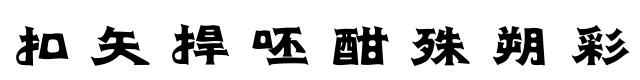}\\
\includegraphics[width=1.5in,height=0.21in]{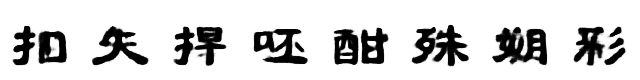}\\
\includegraphics[width=1.5in,height=0.21in]{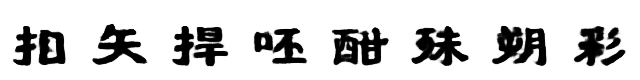}\\
\includegraphics[width=1.5in,height=0.21in]{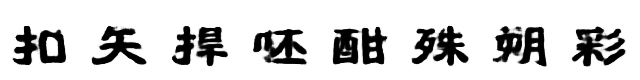}\vspace{3pt}\\
\includegraphics[width=1.5in,height=0.21in]{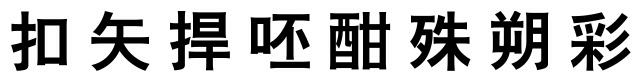}\\
\includegraphics[width=1.5in,height=0.21in]{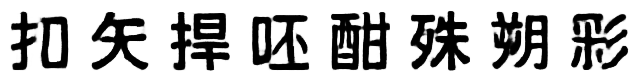}\\
\includegraphics[width=1.5in,height=0.21in]{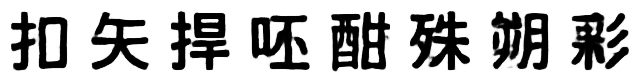}\\
\includegraphics[width=1.5in,height=0.21in]{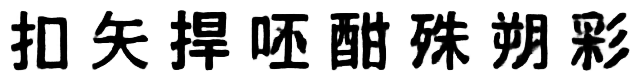}

}
\end{minipage}
}
\caption{The impact of the number of reference images on the generation of images in $D_1$, $D_2$, $D_3$, $D_4$, respectively (from upper left to lower right). TG: Target image, O1: Output for $r$=5, O2: Output for $r$=10, O3: Output for $r$=15. In all cases, $N_{t}$=300k.}
\vspace{-10pt}
\label{fig:ref-num}
\end{figure}

%***********************************com_connection***************************************************
\begin{figure}[!tpb]
\centering
\setlength{\abovecaptionskip}{-2pt}
\hspace{-7pt}
\subfigure{
\begin{minipage}{0.23\textwidth}{\vspace{-5pt}
\begin{minipage}{0.12\textwidth}TG:\end{minipage}
\begin{minipage}{0.15\textwidth}
\includegraphics[width=1.5in,height=0.21in]{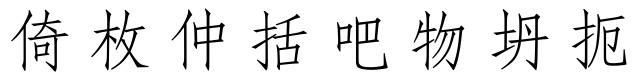}
\end{minipage}\\
\begin{minipage}{0.12\textwidth}O1:\end{minipage}
\begin{minipage}{0.15\textwidth}
\includegraphics[width=1.5in,height=0.21in]{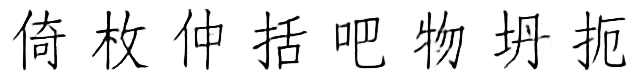}
\end{minipage}\\
\begin{minipage}{0.12\textwidth}O2:\end{minipage}
\begin{minipage}{0.15\textwidth}
\includegraphics[width=1.5in,height=0.21in]{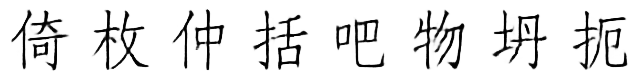}
\end{minipage}
\vspace{3pt}\\
\begin{minipage}{0.12\textwidth}TG:\end{minipage}
\begin{minipage}{0.15\textwidth}
\includegraphics[width=1.5in,height=0.21in]{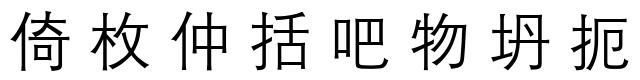}
\end{minipage}\\
\begin{minipage}{0.12\textwidth}O1:\end{minipage}
\begin{minipage}{0.15\textwidth}
\includegraphics[width=1.5in,height=0.21in]{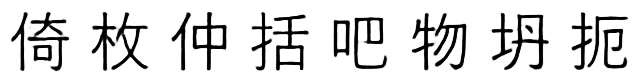}
\end{minipage}\\
\begin{minipage}{0.12\textwidth}O2:\end{minipage}
\begin{minipage}{0.15\textwidth}
\includegraphics[width=1.5in,height=0.21in]{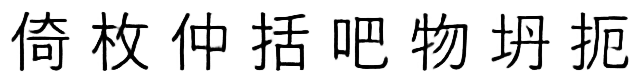}
\end{minipage}
\ruleh
\\
\begin{minipage}{0.12\textwidth}TG:\end{minipage}
\begin{minipage}{0.15\textwidth}
\includegraphics[width=1.5in,height=0.21in]{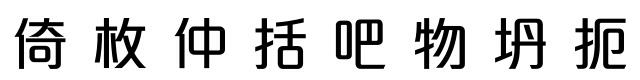}
\end{minipage}\\
\begin{minipage}{0.12\textwidth}O1:\end{minipage}
\begin{minipage}{0.15\textwidth}
\includegraphics[width=1.5in,height=0.21in]{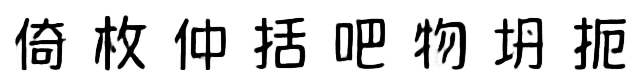}
\end{minipage}\\
\begin{minipage}{0.12\textwidth}O2:\end{minipage}
\begin{minipage}{0.15\textwidth}
\includegraphics[width=1.5in,height=0.21in]{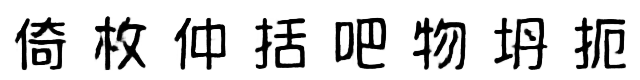}
\end{minipage}
\vspace{3pt}\\
\begin{minipage}{0.12\textwidth}TG:\end{minipage}
\begin{minipage}{0.15\textwidth}
\includegraphics[width=1.5in,height=0.21in]{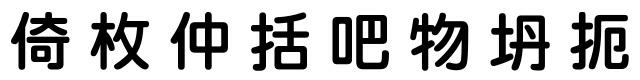}
\end{minipage}\\
\begin{minipage}{0.12\textwidth}O1:\end{minipage}
\begin{minipage}{0.15\textwidth}
\includegraphics[width=1.5in,height=0.21in]{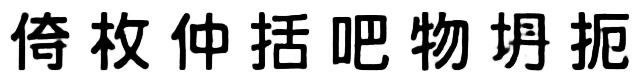}
\end{minipage}\\
\begin{minipage}{0.12\textwidth}O2:\end{minipage}
\begin{minipage}{0.15\textwidth}
\includegraphics[width=1.5in,height=0.21in]{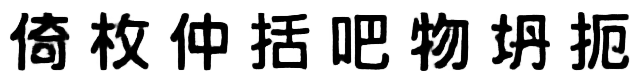}
\end{minipage}
}
\end{minipage}
}
\hspace{1pt}
\rulev
\hspace{-3pt}
\subfigure{
\begin{minipage}{0.21\textwidth}{\vspace{-6pt}
\includegraphics[width=1.5in,height=0.21in]{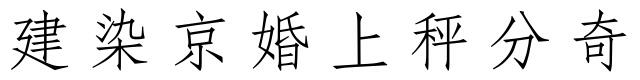}\\
\includegraphics[width=1.5in,height=0.21in]{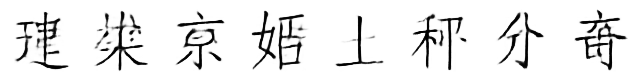}\\
\includegraphics[width=1.5in,height=0.21in]{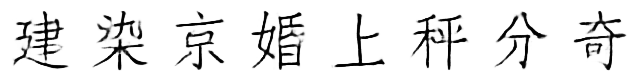}\vspace{3pt}\\
\includegraphics[width=1.5in,height=0.21in]{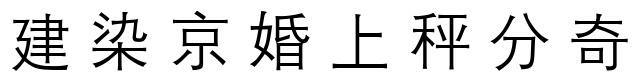}\\
\includegraphics[width=1.5in,height=0.21in]{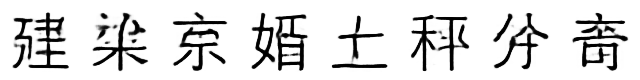}\\
\includegraphics[width=1.5in,height=0.21in]{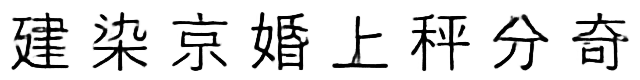}
\ruleh
\vspace{2pt}
\includegraphics[width=1.5in,height=0.21in]{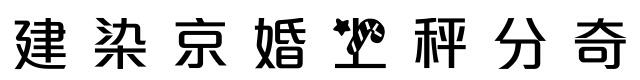}\\
\includegraphics[width=1.5in,height=0.21in]{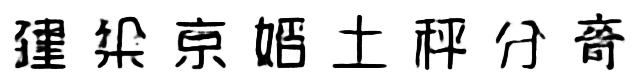}\\
\includegraphics[width=1.5in,height=0.21in]{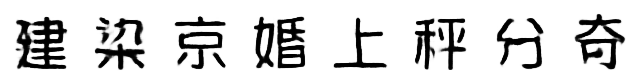}\vspace{3pt}\\
\includegraphics[width=1.5in,height=0.21in]{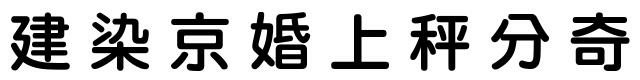}\\
\includegraphics[width=1.5in,height=0.21in]{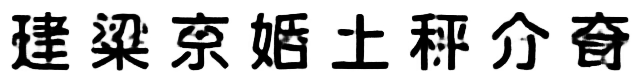}\\
\includegraphics[width=1.5in,height=0.21in]{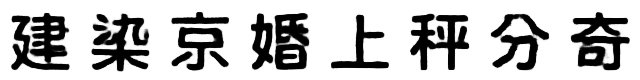}

}
\end{minipage}
}
\caption{The impact of the skip-connection on generation of images in $D_1$, $D_2$, $D_3$, $D_4$, respectively (from upper left to lower right). TG is the target image, O1 and O2 are outputs of models without and with skip-connection. In all cases $N_{t}$=300k, $r$=10.}
\vspace{-15pt}
\label{fig:connection}
\end{figure}

\noindent {\bf Influence of the Reference Set Size}
In addition, we conduct experiments with different number of reference images. Figure~\ref{fig:ref-num} displays the image generation results of $N_{t}$=300k with $r$=5, $r$=10 and $r$=15 respectively. As can be seen from the figure, more reference images lead to better detail generation for characters. Besides, characters generated with $r$=5 are overall okay, meaning that our model can generalize to novel styles using only a few reference images. The generation results of $r$=10 and $r$=15 are close, therefore we take $r$=10 in our other experiments. Intuitively, more reference images supply more information about strokes and styles of characters, making the common points in the reference sets more obvious. Therefore, given $r>1$, our model can achieve co-learning of images with the same style/content. Moreover, with $r>1$ we can learn more images at once which improves the learning efficiency, i.e., if we split the $<$r, r, 1$>$ triplets to be $r^2$ $<$1, 1, 1$>$ triplets, the learning time increases nearly $r^2$ times under the same condition.

\begin{figure}[!t]
\centering
\setlength{\abovecaptionskip}{-2pt}
\hspace{2pt}
\includegraphics[height=1.7in,width=3.4in]{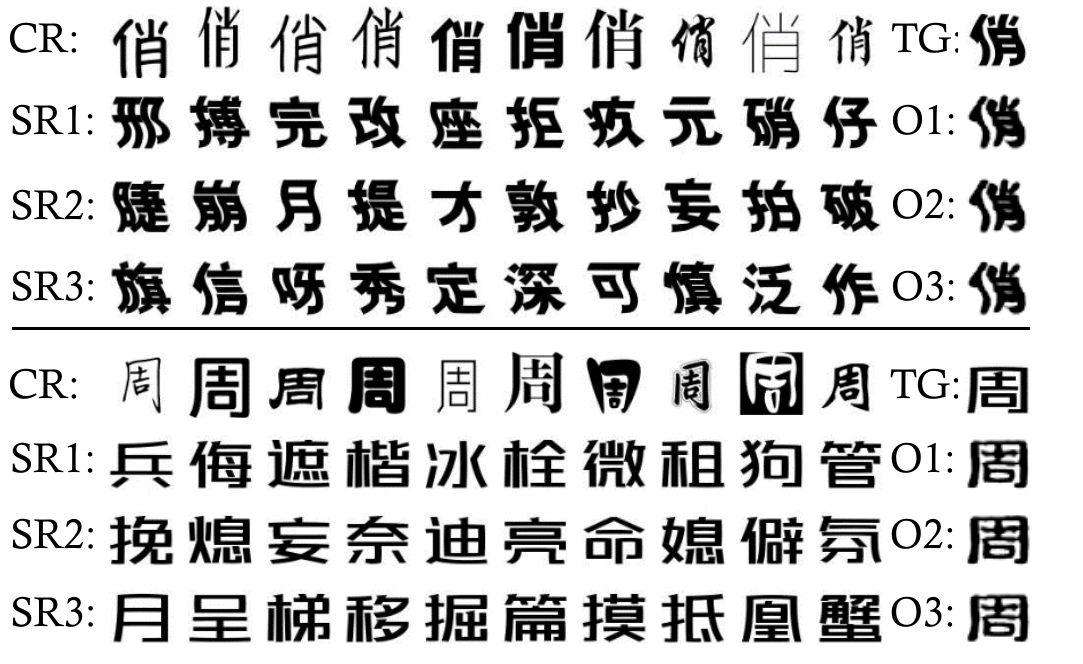}
\caption{Validation of pure style extraction. CR: the content reference set, TG: the target image, O1, O2 and O3 are generated by CR and three different style reference sets SR1, SR2 and SR3.}
\vspace{-10pt}
\label{fig:vali-style}
\end{figure}

%***********************************pure content validation***************************************************

\begin{figure}[!t]
\centering
\setlength{\abovecaptionskip}{-2pt}
\includegraphics[height=1.7in,width=3.3in]{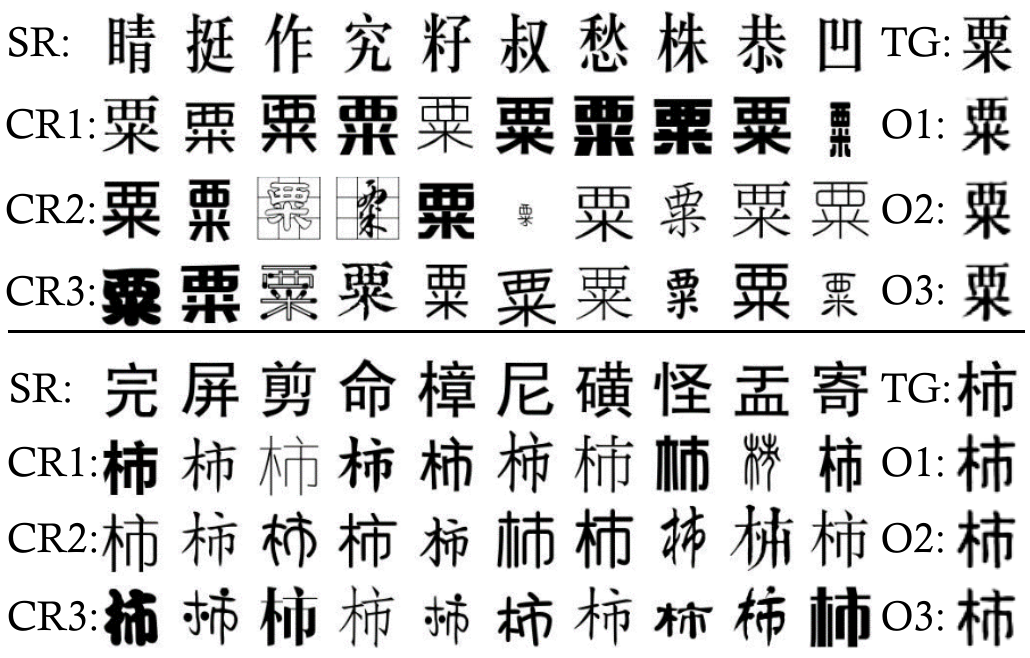}
\caption{Validation of pure content extraction. SR: the style reference set, TG: the target image, O1, O2 and O3 are generated using SR but three different content reference sets CR1, CR2 and CR3.}
\vspace{-10pt}
\label{fig:vali-content}
\end{figure}

%***********************************comparison area2***************************************************
\begin{figure*}[!t]
\centering
\setlength{\abovecaptionskip}{-5pt}
\hspace{-20pt}
\subfigure{
\begin{minipage}{0.42\textwidth}{\vspace{-10pt}
\begin{minipage}{0.17\textwidth}Source:\end{minipage}
\begin{minipage}{0.38\textwidth}
\includegraphics[width=2.4in,height=0.25in]{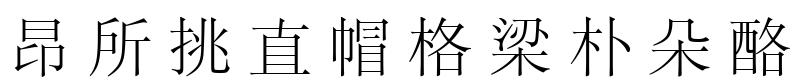}
\end{minipage}\\
\begin{minipage}{0.17\textwidth}Pix2pix:\end{minipage}
\begin{minipage}{0.38\textwidth}
\includegraphics[width=2.4in,height=0.25in]{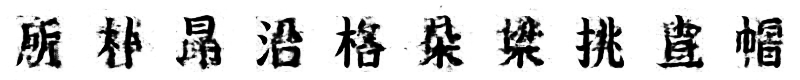}
\end{minipage}\\
\begin{minipage}{0.17\textwidth}AEGN:\end{minipage}
\begin{minipage}{0.38\textwidth}
\includegraphics[width=2.4in,height=0.25in]{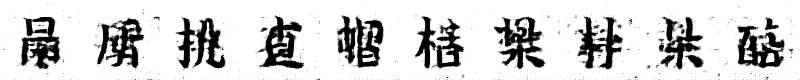}
\end{minipage}\\
\begin{minipage}{0.17\textwidth}Zitozi:\end{minipage}
\begin{minipage}{0.38\textwidth}
\includegraphics[width=2.4in,height=0.25in]{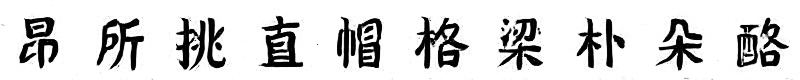}
\end{minipage}\\
\begin{minipage}{0.17\textwidth}C-GAN:\end{minipage}
\begin{minipage}{0.38\textwidth}
\includegraphics[width=2.4in,height=0.25in]{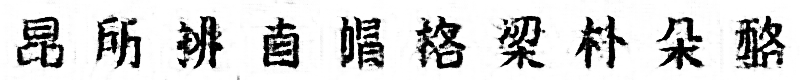}
\end{minipage}\\
\begin{minipage}{0.17\textwidth}EMD:\end{minipage}
\begin{minipage}{0.38\textwidth}
\includegraphics[width=2.4in,height=0.25in]{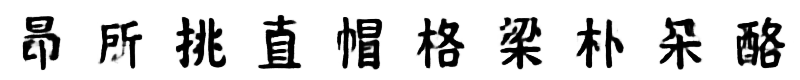}
\end{minipage}\\
\begin{minipage}{0.17\textwidth}Target:\end{minipage}
\begin{minipage}{0.38\textwidth}
\includegraphics[width=2.4in,height=0.25in]{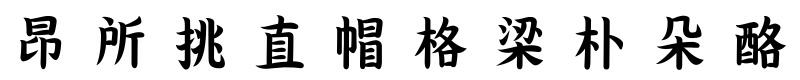}
\end{minipage}
}
\end{minipage}
\rulev
\begin{minipage}{0.35\textwidth}{\vspace{-5pt}
\begin{minipage}{0.38\textwidth}
\includegraphics[width=2.4in,height=0.25in]{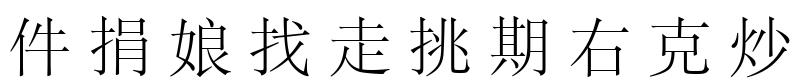}
\end{minipage}\\
\begin{minipage}{0.38\textwidth}
\includegraphics[width=2.4in,height=0.25in]{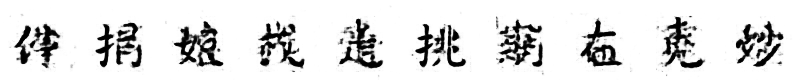}
\end{minipage}\\
\begin{minipage}{0.38\textwidth}
\includegraphics[width=2.4in,height=0.25in]{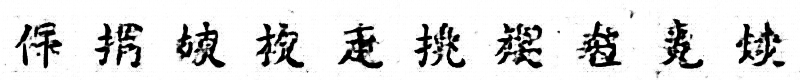}
\end{minipage}\\
\begin{minipage}{0.44\textwidth}
\includegraphics[width=2.4in,height=0.25in]{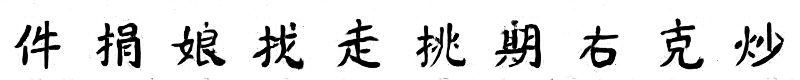}
\end{minipage}\\
\begin{minipage}{0.44\textwidth}
\includegraphics[width=2.4in,height=0.25in]{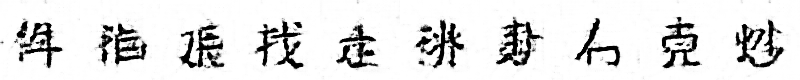}
\end{minipage}\\
\begin{minipage}{0.35\textwidth}
\includegraphics[width=2.4in,height=0.25in]{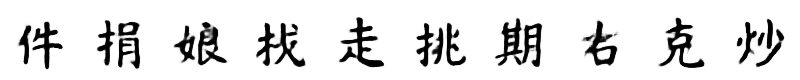}
\end{minipage}\\
\begin{minipage}{0.35\textwidth}
\includegraphics[width=2.4in,height=0.25in]{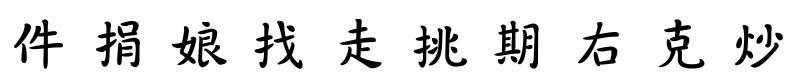}
\end{minipage}\\
}
\end{minipage}
\rulev
\begin{minipage}{0.45\textwidth}{\vspace{-15pt}
\begin{minipage}{0.15\textwidth}L1 loss\end{minipage}
\begin{minipage}{0.15\textwidth}RMSE\end{minipage}
\begin{minipage}{0.15\textwidth}PDAR\end{minipage}\vspace{6pt}\\
\begin{minipage}{0.15\textwidth}0.0105\end{minipage}
\begin{minipage}{0.15\textwidth}0.0202\end{minipage}
\begin{minipage}{0.15\textwidth}0.17\end{minipage}\vspace{6pt}\\
\begin{minipage}{0.15\textwidth}0.0112\end{minipage}
\begin{minipage}{0.15\textwidth}0.0202\end{minipage}
\begin{minipage}{0.15\textwidth}0.3001\end{minipage}\vspace{6pt}\\
\begin{minipage}{0.15\textwidth}0.0091\end{minipage}
\begin{minipage}{0.15\textwidth}\textbf{0.0184}\end{minipage}
\begin{minipage}{0.15\textwidth}0.1659\end{minipage}\vspace{6pt}\\
\begin{minipage}{0.15\textwidth}0.0112\end{minipage}
\begin{minipage}{0.15\textwidth}0.02\end{minipage}
\begin{minipage}{0.15\textwidth}0.3685\end{minipage}\vspace{6pt}\\
\begin{minipage}{0.15\textwidth}\textbf{0.0087}\end{minipage}
\begin{minipage}{0.15\textwidth}\textbf{0.0184}\end{minipage}
\begin{minipage}{0.15\textwidth}\textbf{0.1332}\end{minipage}\\
}
\end{minipage}
}
\caption{Comparison of image generation for known styles and novel contents. Equal number of image pairs with source and target styles are used to train the baselines.}
\label{fig:com-area2}
\vspace{-15pt}
\end{figure*}

\noindent {\bf Effect of the Skip-connection}
To evaluate the effectiveness of the skip-connection during image generation, we compare the results with and without skip-connection in Figure~\ref{fig:connection}. As shown in the figure, images in $D_1$ are generated best, next is $D_3$ and last is $D_2$ and $D_4$, which conforms to the difficulty level and indicates that novel contents are more challenging to extract than novel styles. For known contents, models with and without skip-connection perform closely. But for novel contents, images generated with skip-connection are much better in details. Besides, the model without skip-connection may generate images of novel characters to be similar characters which it has seen before. This is because the structure of novel characters is more challenging to extract and the loss of structure information during down-sampling makes the model generate blurry even wrong characters. However, with content skip-connection, the loss in location and structure information is recaptured by the $\textsl{Decoder}$ network.

\noindent {\bf Validation of Style and Content Separation}
Separating style and content is the key feature of the proposed {\em EMD} model. To validate the clear separation of style and content, we combine one content representation with style representations from a few disjoint style reference sets for one style and check whether the generated images are the same. For better validation, the target images are selected from $D_4$, and the content reference sets and style reference sets are all selected from novel styles and novel contents. Similarly, we combine one style representation with content representations from a few disjoint content reference sets. The results are displayed in Figure~\ref{fig:vali-style} and Figure~\ref{fig:vali-content}, respectively. As shown in Figure~\ref{fig:vali-style}, the generated O1, O2 and O3 are similar although the style reference sets used are quite different, demonstrating that the $\textsl{Style Encoder}$ is able to accurately extract style representations as the only thing the three style reference sets share is the style. Similar results can be found in Figure~\ref{fig:vali-content}, showing that the $\textsl{Content Encoder}$ accurately extracts content representations.

\noindent {\bf Comparison with Baseline Methods}
In the following, we compare our method with the following baselines for character style transfer.
\begin{itemize}
    \item Pix2pix~\cite{isola}: Pix2pix is a conditional GAN based image translation network, which consists of encoder, decoder and a discriminator. It also adopts the skip-connection to connect encoder and decoder. Pix2pix is optimized by L1 distance loss and adversarial loss.
 \item Auto-encoder guided GAN~\cite{lyu}: Auto-encoder guided GAN consists of two encoder-decoder networks, one for image transfer and another acting as an auto-encoder to guide the transfer to learn detailed stroke information.
\item Zi-to-zi~\cite{zitozi}: Zi-to-zi is proposed for Chinese typeface transfer which is based on the encoder-decoder architecture followed by a discriminator. In discriminator, there are two fully connected layers to predict the real/fake and the style category respectively.
\item CycleGAN~\cite{Zhu2017}: CycleGAN consists of two mapping networks which translate images from style A to B and from style B to A, respectively and construct a cycle process. The CycleGAN model is optimized by the adversarial loss and cycle consistency loss.
\end{itemize}

For comparison, we use the font Song as the source font which is simple and commonly used and transfer it to target fonts. Our model is trained with $N_{t}$=300k and $r$=10 and as an average, we use less than 500 images for each style. We compare our method with baselines on generating images with known styles and novel styles, respectively. For novel style, the baselines need to be re-trained from scratch.

%***********************************comparison area4***************************************************
\begin{figure*}[!t]
\centering
\setlength{\abovecaptionskip}{-5pt}
\hspace{-20pt}
\subfigure{
\begin{minipage}{0.45\textwidth}{
\begin{minipage}{0.26\textwidth}Source:\end{minipage}
\begin{minipage}{0.44\textwidth}
\includegraphics[width=2.3in,height=0.25in]{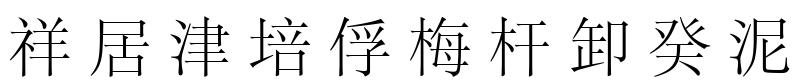}
\end{minipage}\\
\begin{minipage}{0.26\textwidth}Pix2pix-300:\end{minipage}
\begin{minipage}{0.44\textwidth}
\includegraphics[width=2.3in,height=0.25in]{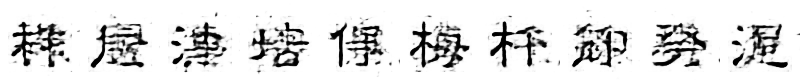}
\end{minipage}\\
\begin{minipage}{0.26\textwidth}Pix2pix-500:\end{minipage}
\begin{minipage}{0.44\textwidth}
\includegraphics[width=2.3in,height=0.25in]{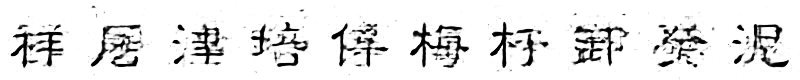}
\end{minipage}\\
\begin{minipage}{0.26\textwidth}Pix2pix-1299:\end{minipage}
\begin{minipage}{0.44\textwidth}
\includegraphics[width=2.3in,height=0.25in]{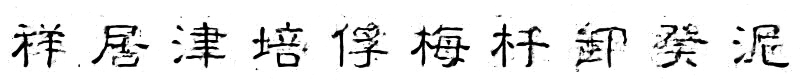}
\end{minipage}\\
\begin{minipage}{0.26\textwidth}AEGN-300:\end{minipage}
\begin{minipage}{0.44\textwidth}
\includegraphics[width=2.3in,height=0.25in]{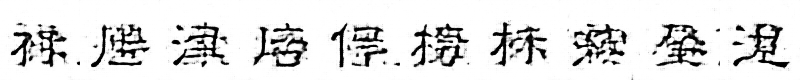}
\end{minipage}\\
\begin{minipage}{0.26\textwidth}AEGN-500:\end{minipage}
\begin{minipage}{0.44\textwidth}
\includegraphics[width=2.3in,height=0.25in]{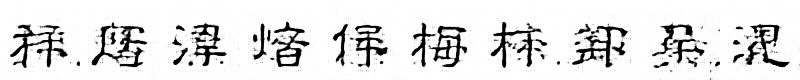}
\end{minipage}\\
\begin{minipage}{0.26\textwidth}AEGN-1299:\end{minipage}
\begin{minipage}{0.44\textwidth}
\includegraphics[width=2.3in,height=0.25in]{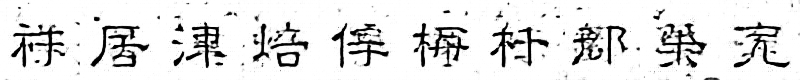}
\end{minipage}\\
\begin{minipage}{0.26\textwidth}Zitozi-300:\end{minipage}
\begin{minipage}{0.44\textwidth}
\includegraphics[width=2.3in,height=0.25in]{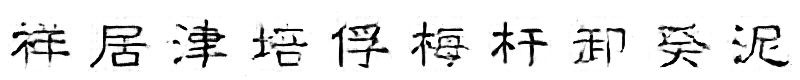}
\end{minipage}\\
\begin{minipage}{0.26\textwidth}Zitozi-500:\end{minipage}
\begin{minipage}{0.44\textwidth}
\includegraphics[width=2.3in,height=0.25in]{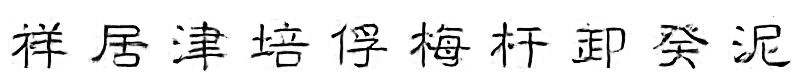}
\end{minipage}\\
\begin{minipage}{0.26\textwidth}Zitozi-1299:\end{minipage}
\begin{minipage}{0.44\textwidth}
\includegraphics[width=2.3in,height=0.25in]{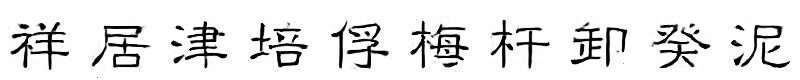}
\end{minipage}\\
\begin{minipage}{0.26\textwidth}C-GAN-300:\end{minipage}
\begin{minipage}{0.44\textwidth}
\includegraphics[width=2.3in,height=0.25in]{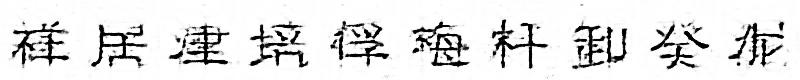}
\end{minipage}\\
\begin{minipage}{0.26\textwidth}C-GAN-500:\end{minipage}
\begin{minipage}{0.44\textwidth}
\includegraphics[width=2.3in,height=0.25in]{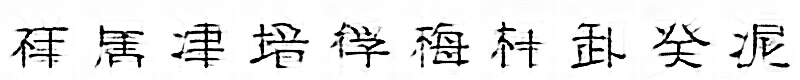}
\end{minipage}\\
\begin{minipage}{0.26\textwidth}C-GAN-1299:\end{minipage}
\begin{minipage}{0.44\textwidth}
\includegraphics[width=2.3in,height=0.25in]{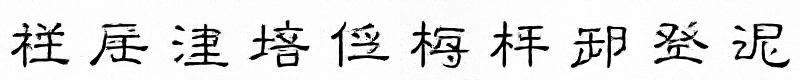}
\end{minipage}\\
\begin{minipage}{0.26\textwidth}EMD-10:\end{minipage}
\begin{minipage}{0.44\textwidth}
\includegraphics[width=2.3in,height=0.25in]{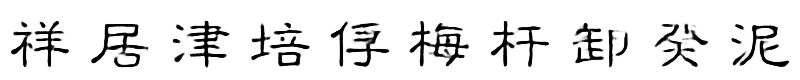}
\end{minipage}\\
\begin{minipage}{0.26\textwidth}Target:\end{minipage}
\begin{minipage}{0.44\textwidth}
\includegraphics[width=2.3in,height=0.25in]{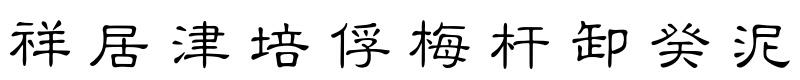}
\end{minipage}\\
}
\end{minipage}
\rulev
\begin{minipage}{0.33\textwidth}{\vspace{-5pt}
\begin{minipage}{0.44\textwidth}
\includegraphics[width=2.3in,height=0.25in]{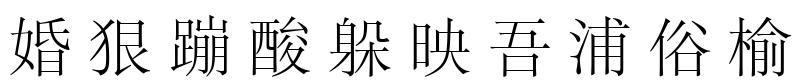}
\end{minipage}\\
\begin{minipage}{0.44\textwidth}
\includegraphics[width=2.3in,height=0.25in]{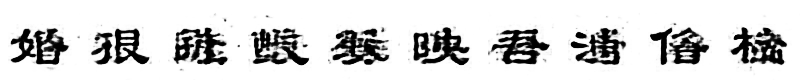}
\end{minipage}\\
\begin{minipage}{0.44\textwidth}
\includegraphics[width=2.3in,height=0.25in]{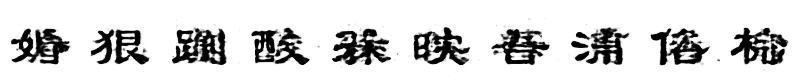}
\end{minipage}\\
\begin{minipage}{0.44\textwidth}
\includegraphics[width=2.3in,height=0.25in]{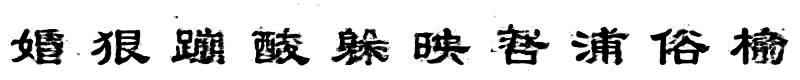}
\end{minipage}\\
\begin{minipage}{0.44\textwidth}
\includegraphics[width=2.3in,height=0.25in]{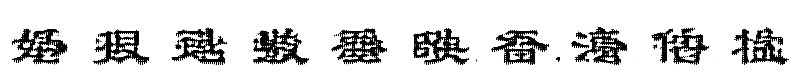}
\end{minipage}\\
\begin{minipage}{0.44\textwidth}
\includegraphics[width=2.3in,height=0.25in]{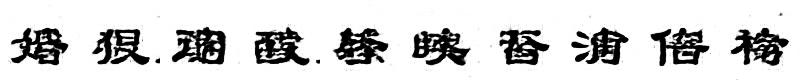}
\end{minipage}\\
\begin{minipage}{0.44\textwidth}
\includegraphics[width=2.3in,height=0.25in]{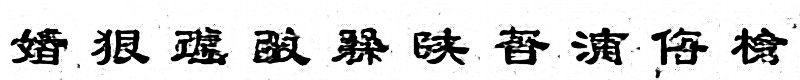}
\end{minipage}\\
\begin{minipage}{0.44\textwidth}
\includegraphics[width=2.3in,height=0.25in]{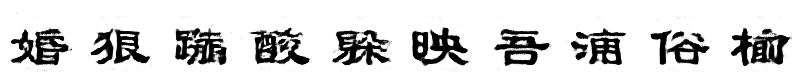}
\end{minipage}\\
\begin{minipage}{0.44\textwidth}
\includegraphics[width=2.3in,height=0.25in]{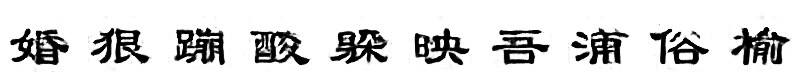}
\end{minipage}\\
\begin{minipage}{0.44\textwidth}
\includegraphics[width=2.3in,height=0.25in]{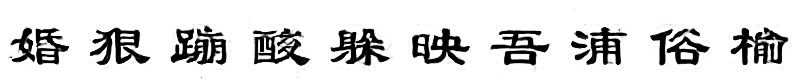}
\end{minipage}\\
\begin{minipage}{0.44\textwidth}
\includegraphics[width=2.3in,height=0.25in]{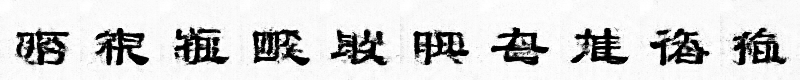}
\end{minipage}\\
\begin{minipage}{0.44\textwidth}
\includegraphics[width=2.3in,height=0.25in]{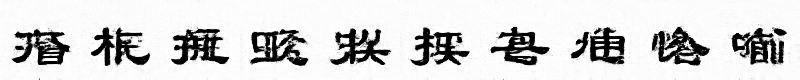}
\end{minipage}\\
\begin{minipage}{0.44\textwidth}
\includegraphics[width=2.3in,height=0.25in]{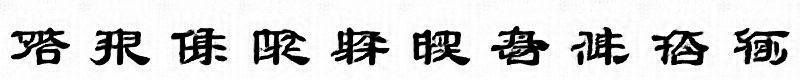}
\end{minipage}\\
\begin{minipage}{0.44\textwidth}
\includegraphics[width=2.3in,height=0.25in]{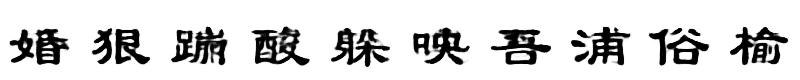}
\end{minipage}\\
\begin{minipage}{0.44\textwidth}
\includegraphics[width=2.3in,height=0.25in]{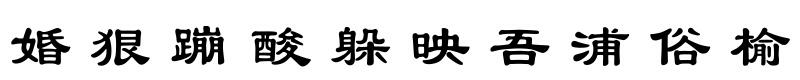}
\end{minipage}
}
\end{minipage}
\rulev
\begin{minipage}{0.40\textwidth}{\vspace{-20pt}
\begin{minipage}{0.15\textwidth}L1 loss\end{minipage}
\begin{minipage}{0.15\textwidth}RMSE\end{minipage}
\begin{minipage}{0.15\textwidth}PDAR\end{minipage}\vspace{6pt}\\
\begin{minipage}{0.15\textwidth}0.0109\end{minipage}
\begin{minipage}{0.15\textwidth}0.0206\end{minipage}
\begin{minipage}{0.15\textwidth}0.1798\end{minipage}\vspace{6pt}\\
\begin{minipage}{0.15\textwidth}0.0106\end{minipage}
\begin{minipage}{0.15\textwidth}0.0202\end{minipage}
\begin{minipage}{0.15\textwidth}0.1765\end{minipage}\vspace{6pt}\\
\begin{minipage}{0.15\textwidth}0.01\end{minipage}
\begin{minipage}{0.15\textwidth}0.0196\end{minipage}
\begin{minipage}{0.15\textwidth}0.1531\end{minipage}\vspace{6pt}\\
\begin{minipage}{0.15\textwidth}0.0117\end{minipage}
\begin{minipage}{0.15\textwidth}0.02\end{minipage}
\begin{minipage}{0.15\textwidth}0.3951\end{minipage}\vspace{6pt}\\
\begin{minipage}{0.15\textwidth}0.0108\end{minipage}
\begin{minipage}{0.15\textwidth}0.02\end{minipage}
\begin{minipage}{0.15\textwidth}0.2727\end{minipage}\vspace{6pt}\\
\begin{minipage}{0.15\textwidth}0.0105\end{minipage}
\begin{minipage}{0.15\textwidth}0.0196\end{minipage}
\begin{minipage}{0.15\textwidth}0.26\end{minipage}\vspace{6pt}\\
\begin{minipage}{0.15\textwidth}0.0091\end{minipage}
\begin{minipage}{0.15\textwidth}0.0187\end{minipage}
\begin{minipage}{0.15\textwidth}0.1612\end{minipage}\vspace{6pt}\\
\begin{minipage}{0.15\textwidth}\textbf{0.009}\end{minipage}
\begin{minipage}{0.15\textwidth}0.0185\end{minipage}
\begin{minipage}{0.15\textwidth}0.1599\end{minipage}\vspace{6pt}\\
\begin{minipage}{0.15\textwidth}\textbf{0.009}\end{minipage}
\begin{minipage}{0.15\textwidth}\textbf{0.0183}\end{minipage}
\begin{minipage}{0.15\textwidth}0.1624\end{minipage}\vspace{6pt}\\
\begin{minipage}{0.15\textwidth}0.0143\end{minipage}
\begin{minipage}{0.15\textwidth}0.0215\end{minipage}
\begin{minipage}{0.15\textwidth}0.5479\end{minipage}\vspace{6pt}\\
\begin{minipage}{0.15\textwidth}0.0126\end{minipage}
\begin{minipage}{0.15\textwidth}0.0203\end{minipage}
\begin{minipage}{0.15\textwidth}0.4925\end{minipage}\vspace{6pt}\\
\begin{minipage}{0.15\textwidth}0.0128\end{minipage}
\begin{minipage}{0.15\textwidth}0.0203\end{minipage}
\begin{minipage}{0.15\textwidth}0.4885\end{minipage}\vspace{6pt}\\
\begin{minipage}{0.15\textwidth}\textbf{0.009}\end{minipage}
\begin{minipage}{0.15\textwidth}0.0186\end{minipage}
\begin{minipage}{0.15\textwidth}\textbf{0.1389}\end{minipage}
}

\end{minipage}
}
\caption{Comparison of image generation for novel styles and contents given $r$=10. The baseline methods are trained with 300, 500, 1299 image pairs respectively.}
\vspace{-15pt}
\label{fig:com-area4}
\end{figure*}

\textbf{Known styles as target style.} Taking known styles as the target style, baselines are trained using the same number of paired images as the images our model used for the target style. The results are displayed in Figure~\ref{fig:com-area2} where CycleGAN is denoted as C-GAN for simplicity. We can observe that for known styles and novel contents, our method performs much better than pix2pix, AEGN and CycleGAN and close to or even slightly better than zi-to-zi. This is because pix2pix and AEGN usually need more samples to learn a style~\cite{lyu}. CycleGAN performs poorly and only generates part of characters or some strokes, possibly because it learns the domain mappings without the domain knowledge. Zitozi performs well since it learns multiple styles at the same time and the contrast among different styles helps the model better learn styles.

For quantitative analysis, we calculate the L1 loss, Root Mean Square Error (RMSE) and the Pixel Disagreement Ratio (PDAR)~\cite{Zhu2017} between the generated images and the target images. PDAR is the number of pixels with different values in the two images divided by the total image size after image binaryzation. We conduct experiments for 10 randomly sampled styles and the average results are displayed at the last three columns in Figure~\ref{fig:com-area2} and the best performance is bold. We can observe that our method performs best and achieves the lowest L1 loss, RMSE and PDAR.

\textbf{Novel styles as target style.} Taking novel styles as the target style, we test our model to generate images of novel styles and contents given $r$=10 style/content reference images without retraining. As for baselines, retraining is needed. Here, we conduct two experiments for baselines. One is that we first pretrain a model for each baseline method using the training set our method used and then fine-tune the pretrained model with the same 10 reference images as our method used. The results show that all baseline methods preform poorly and it is unfeasible to learn a style by fine-tuning on only 10 reference images. Thus, we omit the experiment results here. The other setting is training the baseline model from scratch. Since it is unrealistic to train baseline models with only 10 samples, we train them using 300, 500, 1299 images of the target style respectively. Here we use 1299 is because the number of train contents is 1299 in our data set. The results are presented in Figure~\ref{fig:com-area4}. As shown in the figure, the proposed {\em EMD} model can generalize to novel styles from only 10 style reference images but other methods need to be retrained with more samples. The pix2pix, AEGN and CycleGAN perform worst even trained with all 1299 training images, which demonstrates that these three methods are not effective for character style transfer especially when the training data are limited. With only 10 style reference images, our model performs better than zi-to-zi-300 namely zi-to-zi model learned with 300 examples for each style, close to zi-to-zi-500 and a little worse than zi-to-zi-1299. This may be because zi-to-zi learns multiple styles at the same time and learning with style contrast helps model learning better.

The quantitative comparison results for L1 loss, RMSE and PDAR are shown at the last three columns of Figure~\ref{fig:com-area4}. Although given only 10 style reference images, our method performs better than all pix2pix, AEGN and CycleGAN models and zi-to-zi-300, and close to zi-to-zi-500 and zi-to-zi-1299, which demonstrates the effectiveness of our method.

In conclusion, these baseline methods require many images of source styles and target styles, which may be difficult to collect. Besides, the learned baseline model can only transfer styles appearing in train set and they have to be retrained for new styles. But our method can generalize to novel styles given only a few reference images. In addition, baseline models can only use images of target styles. However, since the proposed {\em EMD} model learns feature representations instead of transformation among specific styles, it can leverage images of any styles and make the most of existing data.

\subsection{Neural Style Transfer}
%We also validate the proposed {\em EMD} framework on neural style transfer task. First, we introduce the implementation details and then we introduce the baseline methods compared with {\em EMD}. Finally, we analyze the experimental results.

%################### comparison results#########################################
\begin{figure*}[!tbp]
\centering
\vspace{-20pt}
\subfigure{
\includegraphics[height=5.0in,width=7.2in]{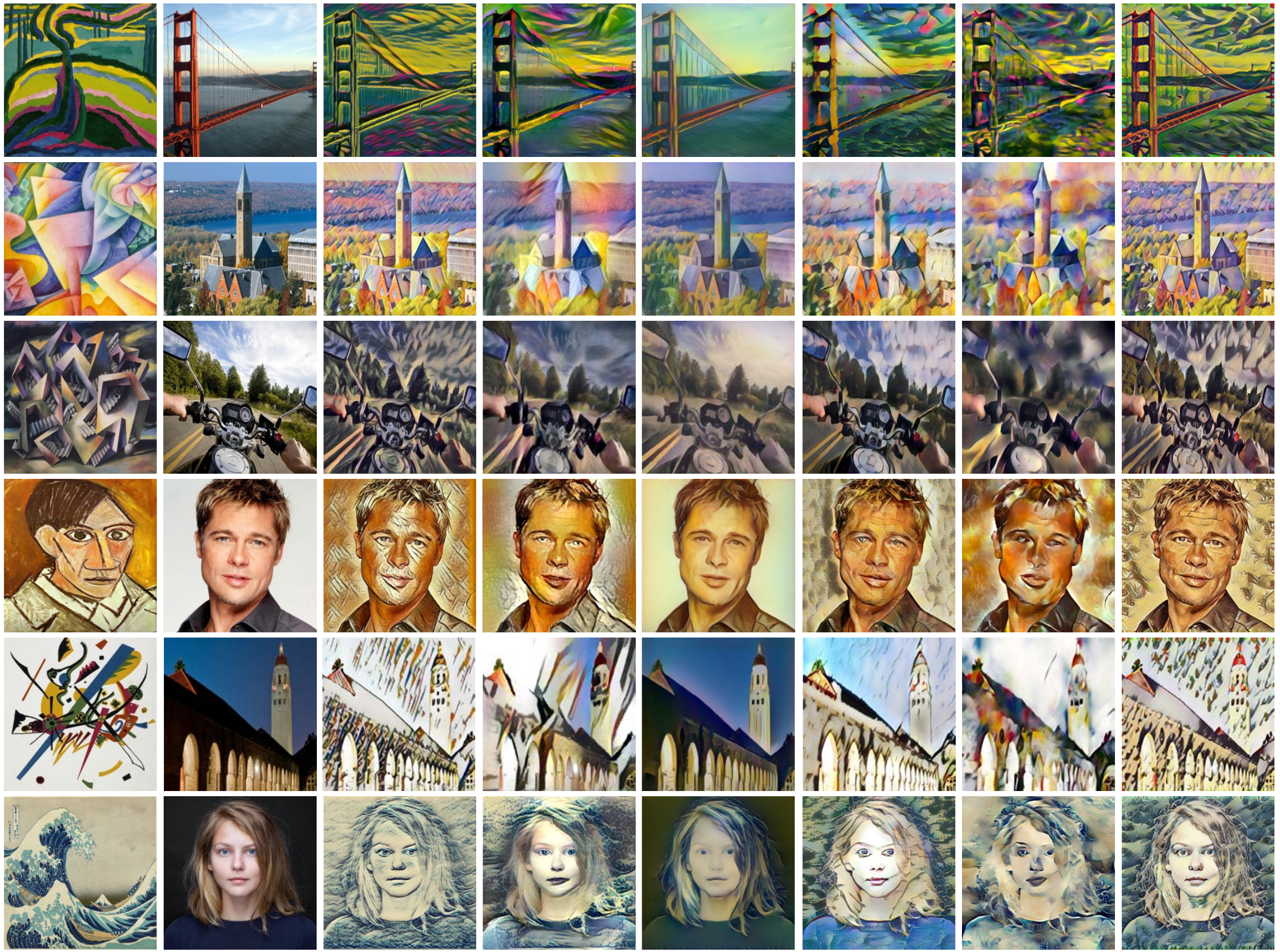}
}
\\
\vspace{-10pt}
\subfigure{
\hspace{-3pt}
\includegraphics[height=3.4in,width=7.2in]{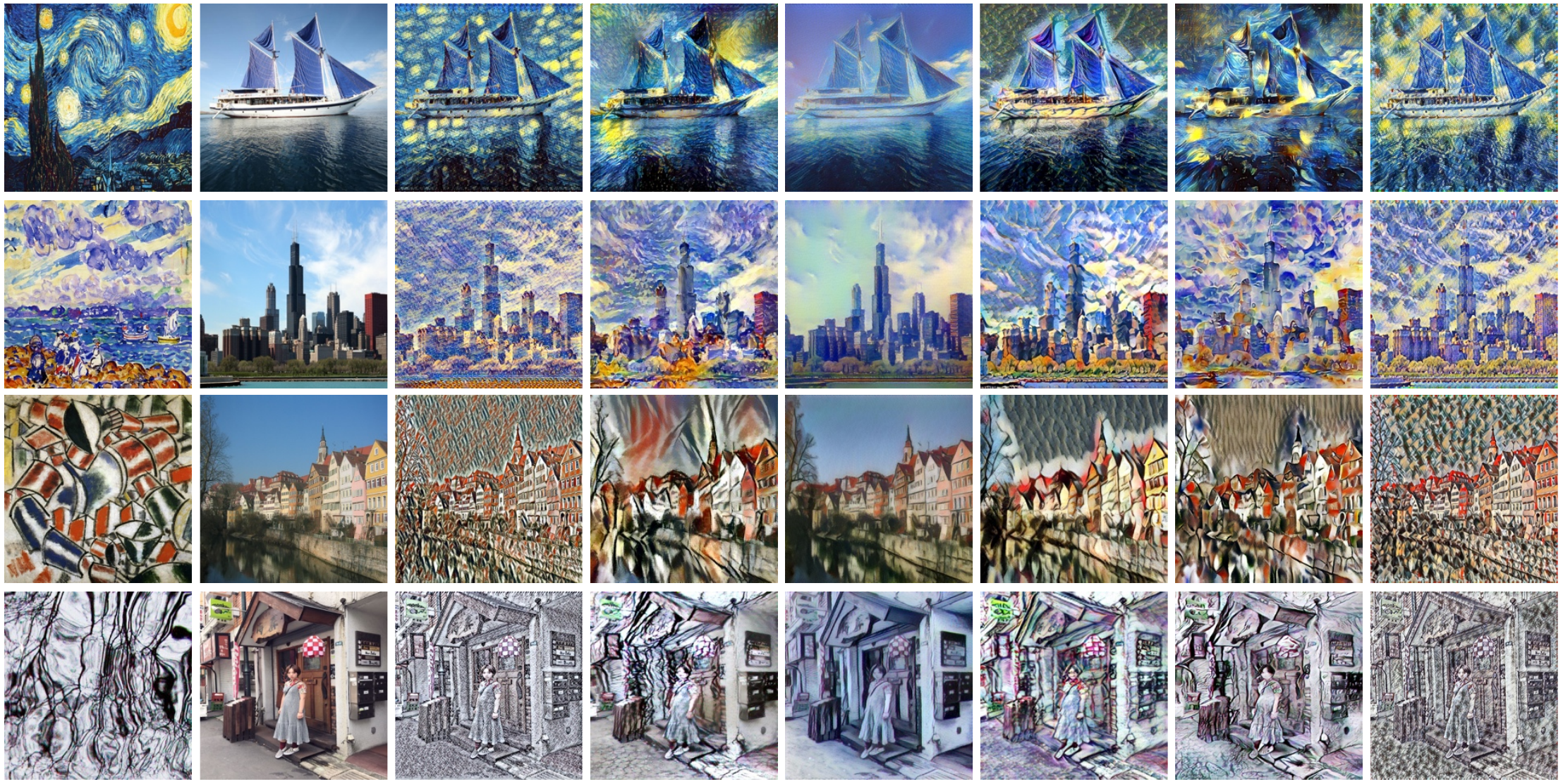}
}
\\
\subfigure{
\begin{minipage}{0.12\textwidth}\centering Style\end{minipage}
\begin{minipage}{0.12\textwidth}\centering Content\end{minipage}
\begin{minipage}{0.12\textwidth}\centering TextureNet\cite{ulyanov2017improved}\end{minipage}
\begin{minipage}{0.12\textwidth}\centering Opt-based\cite{Gatys}\end{minipage}
\begin{minipage}{0.12\textwidth}\centering Patch-based\cite{chen2016fast}\end{minipage}
\begin{minipage}{0.12\textwidth}\centering AdaIn\cite{Huang_2017_ICCV}\end{minipage}
\begin{minipage}{0.12\textwidth}\centering Universal\cite{li2017universal}\end{minipage}
\begin{minipage}{0.12\textwidth}\centering EMD\end{minipage}
}
\caption{The comparison results for neural style transfer.}
\label{fig:neural1}
\vspace{-5pt}
\end{figure*}

%################### more results#########################################
\begin{figure*}[!tbp]
\setlength{\abovecaptionskip}{-2pt}
% \vspace{-20pt}
\centering
\subfigure{
\includegraphics[height=6.0in,width=7.2in]{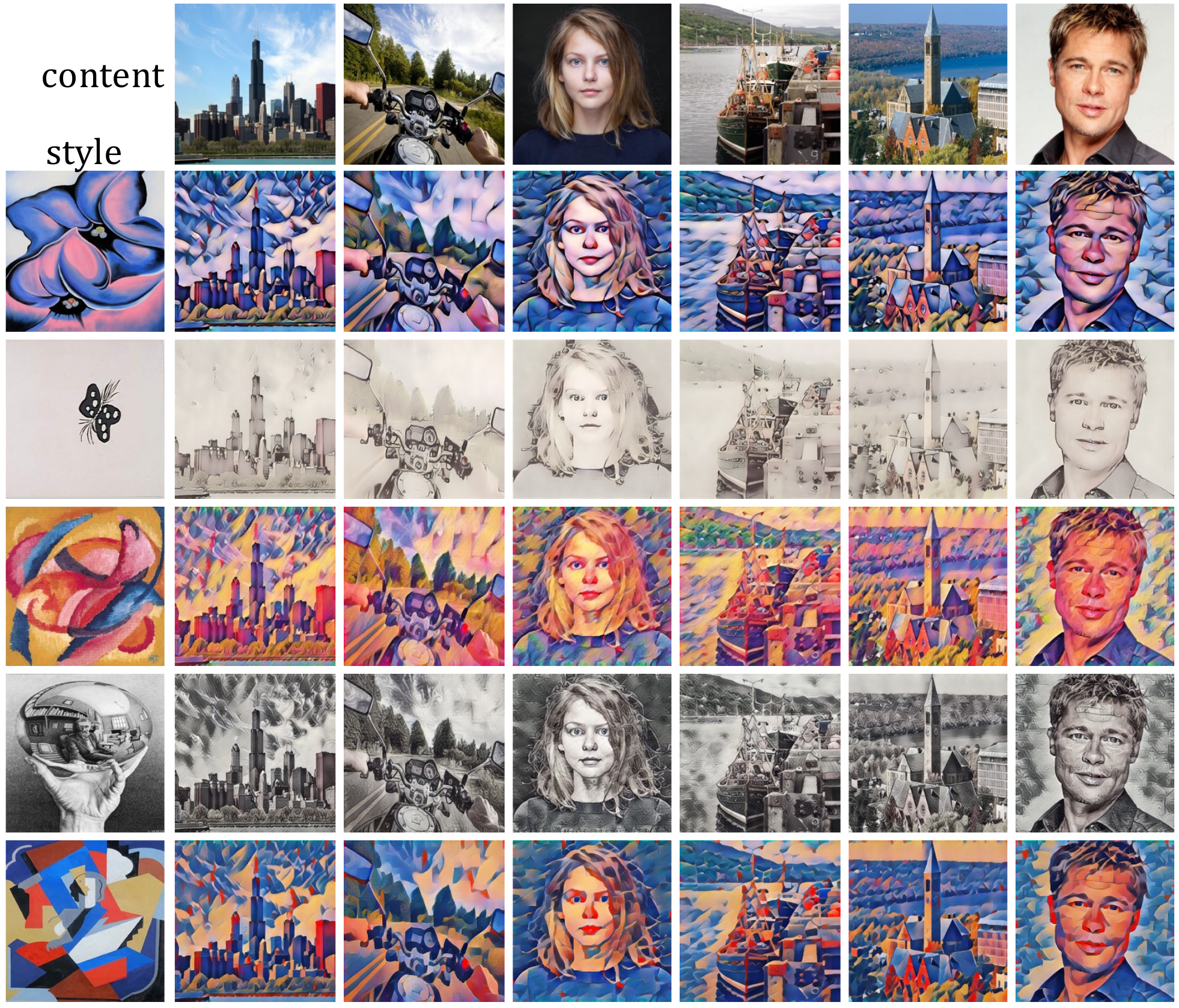}
}
\\
\vspace{-10pt}
\subfigure{
\includegraphics[height=1.0in,width=7.2in]{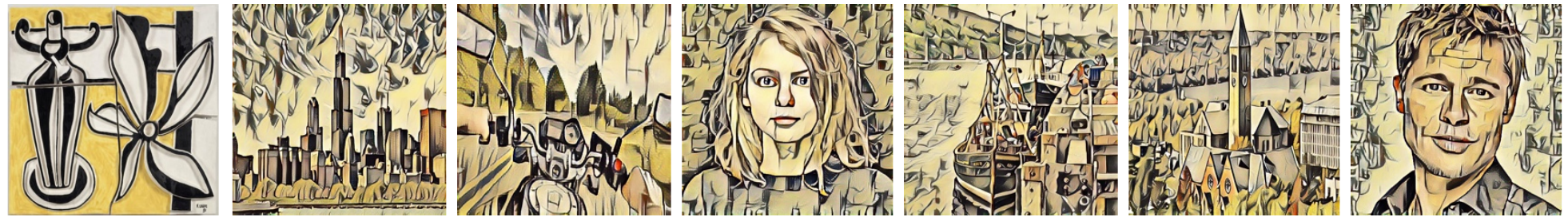}
}
\caption{More experimental results for neural style transfer.}
\label{fig:neural2}
\vspace{-10pt}
\end{figure*}

\subsubsection{Implementation Details}
Following previous studies~\cite{Huang_2017_ICCV,johnson}, we use the MS-COCO dataset~\cite{lin2014microsoft} as the content images and a dataset of paintings mainly collected from WikiArt~\cite{wikiart} as the style images. Each dataset contains roughly 80,000 training examples. The model is trained using the Adam optimizer with the learning rate of 0.0001. The batch size is set to be 8 style-content pairs. We compute the style loss using the ${relu1\_2, relu2\_2, relu3\_3, relu4\_3}$ layers of VGG-19 and the content loss using the ${relu4\_1}$ layer. We set $\lambda_{c}$=1, $\lambda_{s}$=5 and $\lambda_{tv}$=1e-5. During training, we first resize the smallest dimension of both images to 512 while preserving the aspect ratio, then randomly crop regions of size
256$\times$256. Since the size of the fully connected layer in $\textsl{Style Encoder}$ is only related to the filter numbers, our model can be applied to style/content images of any size during testing.

%################### style-content trade-off #########################################
\begin{figure*}[!tbp]
\setlength{\abovecaptionskip}{-2pt}
\centering
\hspace{-10pt}
\subfigure{
\includegraphics[height=1.2in,width=7.2in]{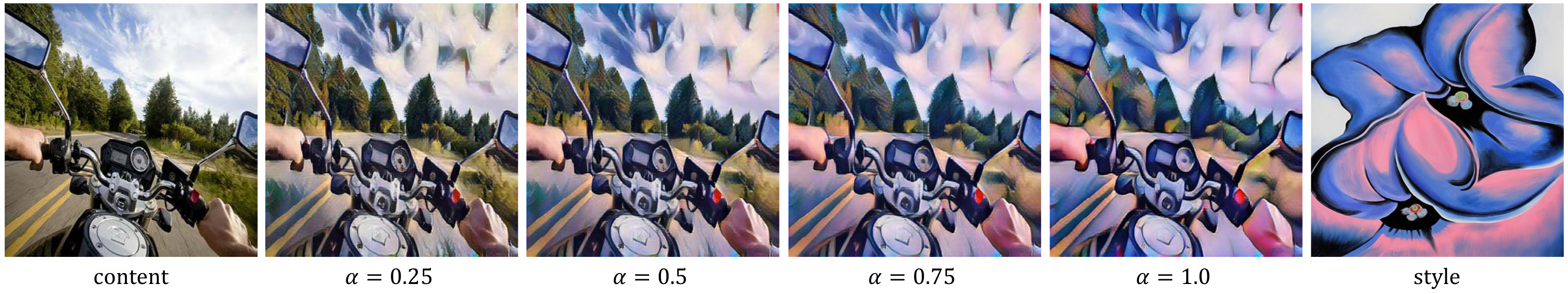}
}
\\
\vspace{-10pt}
\subfigure{
\includegraphics[height=1.2in,width=7.2in]{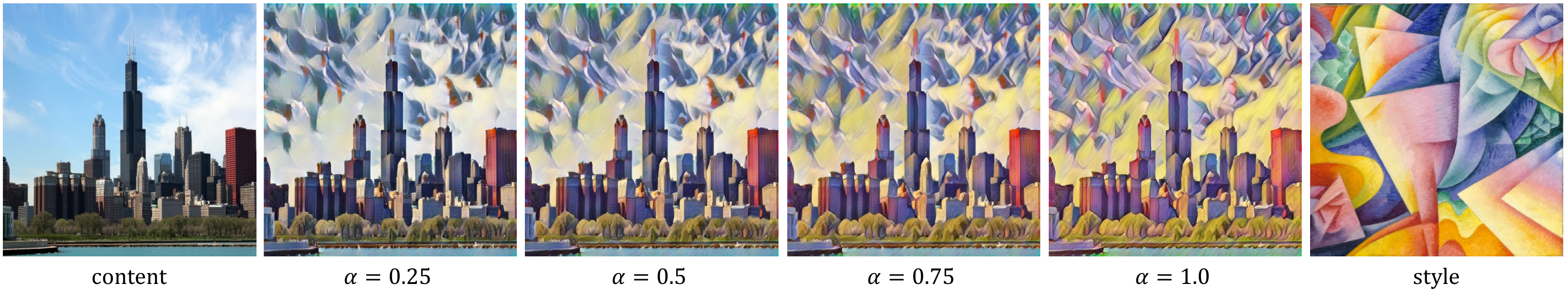}
}
\caption{Examples of style-content trade-off.}
\label{fig:trade-off}
\vspace{-5pt}
\end{figure*}

%################### interpolation#########################################
\begin{figure*}[!tbp]
\setlength{\abovecaptionskip}{-2pt}
\centering
\hspace{-10pt}
\subfigure{
\includegraphics[height=1.1in,width=7.2in]{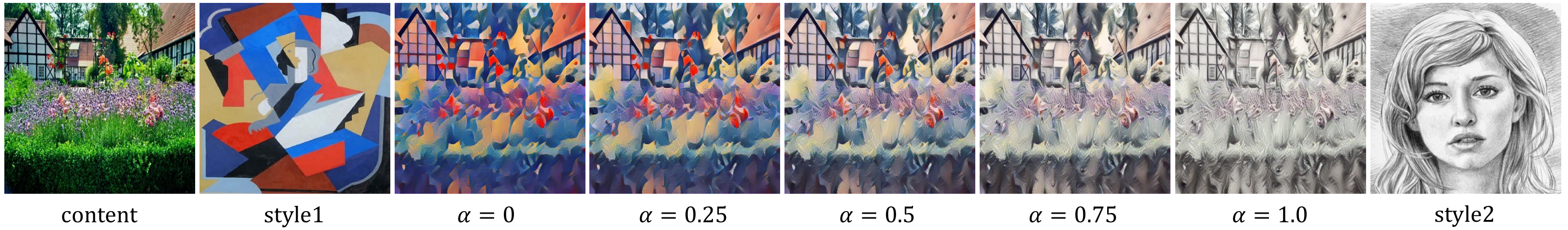}
}
\caption{Examples of style interpolation.}
\label{fig:interpolation}
\vspace{-5pt}
\end{figure*}

\subsubsection{Comparison Methods}
We compare the proposed neural style transfer model with the following three types of baseline methods:
\begin{itemize}
 \item Fast but not flexible Per-Style-Per-Model method, which is restricted to a single style and can not be generalized to new styles. Here we use the state-of-the-art method TextureNet~\cite{ulyanov2017improved} as an example. TextureNet is mainly a generator which takes a noise variable $z$ and a content reference image as the inputs and generates the image with target style/content.
 \item Flexible but slow optimization based method~\cite{Gatys}, which optimizes one noise image to be with target style and content iteratively with the help of a pretrained VGG network.
 \item Flexible and fast Arbitrary-Style-Per-Model method, which can achieve arbitrary style transfer with no need for retraining. In this study, we compare with the following three methods:
     \begin{itemize}
         \item Patch-based~\cite{chen2016fast}: Patch-based method conducts style transfer by swapping each content feature patch with the nearest style patch. The network consists of a convolution network, an inverse network and a style swap layer.
         \item AdaIn~\cite{Huang_2017_ICCV}: AdaIn is based on the Adaptive Instance Normalization and the network of AdaIn consists of an encoder, a decoder and an Adaptive Instance Normalization layer, where the encoder is fixed as the first few layers of VGG-19.
         \item Universal~\cite{li2017universal}: Universal is designed based on the whitening and coloring transformation which is embedded in a series of pretrained encoder-decoder image reconstruction networks.
    \end{itemize}
\end{itemize}
Among the above baseline methods, the TextureNet is more impressive in transfer quality than the other four baseline methods, therefore, we take it as a benchmark. The results of these baseline methods are all obtained by running their released code with the default configurations.

\subsubsection{Experimental Results}
%In this subsection, we first display the comparison results with baseline methods and then we display the results for style-content trade-off and style interpolation.

\noindent {\bf Comparison with Baseline Methods}
As can be seen from Figure~\ref{fig:neural1}, the proposed method performs better than other arbitrary style transfer methods but a little worse than TextureNet. It is worth noting that TextureNet is trained separately for each style but none of the presented styles are observed by our model during training. This is acceptable due to the trade-off between flexibility and transfer quality. Patch-based method performs poorly. It can not capture some styles when lots of content patches are swapped with style patches lack of style elements. AdaIn performs well on most styles but the generated images are a little blurry in details. It performs not so well for some complicated styles. Universal replaces the training process with a series of transformations but it is not effective at producing sharp details and fine strokes. Figure~\ref{fig:neural2} displays more style transfer results of our proposed method, which demonstrate that the proposed {\em EMD} framework can be generalized to arbitrary new styles without the need for model retraining.

\noindent {\bf Style-content Trade-off}
During training, we can control the degree of style transfer by adjusting the weight $\lambda_s$ in loss function. When testing, our method also allows the style-content trade-off by adjusting the amount of style information mixed with the content feature.
With $\textsl{Style Encoder}$, we can obtain the original style of the content image, and then we mix the content feature with the style which is the weighted combination of styles from the content image and the style image
\begin{equation}
    \hat{F} = \frac{F_{con}- \mu(F_{con})}{\sigma(F_{con})} \sigma_\textsl{new} + \mu_\textrm{new},
    \label{eq:trade-off}
\end{equation}
where $F_{con}$ is the feature map of content image and
\begin{equation}
    \mu_\textrm{new} = (1-\alpha)\mu_{con} + \alpha \mu_{sty},
\end{equation}
\begin{equation}
    \sigma_\textrm{new} = (1-\alpha)\sigma_{con} + \alpha \sigma_{sty},
\end{equation}
where ($\mu_{con}$, $\sigma_{con}$) and ($\mu_{sty}$, $\sigma_{sty}$) are the learned statistical information of the content image and the style image, respectively. By adjusting the weight $\alpha$, the $\textsl{Decoder}$ generates images gradually changing from the original style to the target style. When $\alpha=0$, the $\textsl{Decoder}$ tries to reconstruct the content image and when $\alpha=1.0$, the $\textsl{Decoder}$ outputs the most stylized image. As shown in Figure~\ref{fig:trade-off}, the stylized image changes from slightly stylized to the most stylized with increasing $\alpha$.

\noindent {\bf Style Interpolation}
Similarly, our method can also be applied for interpolation between two styles, which is achieved by setting $\mu_\textrm{new} = (1-\alpha) \mu_{sty1}+\alpha \mu_{sty2}$ and $\sigma_\textrm{new} = (1-\alpha) \sigma_{sty1} + \alpha \sigma_{sty2}$ in Eq.~\ref{eq:trade-off}. An example is presented in Figure~\ref{fig:interpolation}. When $\alpha=0$ and $\alpha=1$, style 1 and style 2 are used for the transfer, respectively. When $0<\alpha<1$, an interpolation between the two styles are used for the transfer.

\section{Conclusion and Future Work}
In this paper, we propose a unified style transfer framework {\em EMD} for both character typeface transfer and neural style transfer, which enables the transfer models generalizable to new styles and contents given a few reference images. The main idea is that from these reference images, the $\textsl{Style Encoder}$ and $\textsl{Content Encoder}$ extract style and content representations, respectively. Then the extracted style and content representations are mixed by a $\textsl{Mixer}$ and finally fed into the $\textsl{Decoder}$ to generate images with target styles and contents. This learning framework allows simultaneous style transfer among multiple styles and can be deemed as a special `multi-task' learning scenario. Then the learned encoders, mixer and decoder will be taken as the shared knowledge and transferred to new styles and contents. Under this framework, we design two individual networks for character typeface transfer and neural style transfer tasks. Extensive experimental results on these two tasks demonstrate its effectiveness.

In our study, the learning process consists of a series of image generation tasks and we try to learn a model which can generalize to new but related tasks by learning a high-level strategy, namely learning the style and content representations. This resembles to ``learning-to-learn" program. In the future, we will explore more about ``learning-to-learn" and integrate it with our framework.

\section*{Acknowledgment}
The work is partially supported by the High Technology Research and Development Program of China 2015AA015801, NSFC 61521062, STCSM 18DZ2270700.

\ifCLASSOPTIONcaptionsoff
  \newpage
\fi

{
\bibliographystyle{ieee}
\bibliography{style}

\begin{thebibliography}{10}\itemsep=-1pt

\bibitem{rewrite}
Rewrite.
\newblock https://github.com/kaonashi-tyc/Rewrite.

\bibitem{zitozi}
Zi-to-zi.
\newblock https://kaonashi-tyc.github.io/2017/04/06/zi2zi.html.

\bibitem{wikiart}
Painter by numbers, wikiart, 2016.
\newblock https://www.kaggle.com/c/painter-by-numbers.

\bibitem{azadi2017multi}
S.~Azadi, M.~Fisher, V.~Kim, Z.~Wang, E.~Shechtman, and T.~Darrell.
\newblock Multi-content gan for few-shot font style transfer.
\newblock 2018.

\bibitem{Bousmalis}
K.~Bousmalis, N.~Silberman, D.~Dohan, D.~Erhan, and D.~Krishnan.
\newblock Unsupervised pixel-level domain adaptation with generative
  adversarial networks.
\newblock In {\em Proceedings of the IEEE Conference on Computer Vision and
  Pattern Recognition}, 2017.

\bibitem{changpinyo}
S.~Changpinyo, W.~Chao, B.~Gong, and F.~Sha.
\newblock Synthesized classifiers for zero-shot learning.
\newblock In {\em Proceedings of the IEEE Conference on Computer Vision and
  Pattern Recognition}, pages 5327--5336, 2016.

\bibitem{chen2017}
D.~Chen, L.~Yuan, J.~Liao, N.~Yu, and G.~Hua.
\newblock Stylebank: An explicit representation for neural image style
  transfer.
\newblock In {\em Proceedings of the IEEE Conference on Computer Vision and
  Pattern Recognition}, 2017.

\bibitem{chen2016fast}
T.~Q. Chen and M.~Schmidt.
\newblock Fast patch-based style transfer of arbitrary style.
\newblock {\em arXiv preprint arXiv:1612.04337}, 2016.

\bibitem{dumoulin2016learned}
V.~Dumoulin, J.~Shlens, and M.~Kudlur.
\newblock A learned representation for artistic style.
\newblock In {\em Proceedings of the International Conference on Learning
  Representations}, 2017.

\bibitem{frome}
A.~Frome, G.~Corrado, J.~Shlens, S.~Bengio, J.~Dean, T.~Mikolov, et~al.
\newblock Devise: A deep visual-semantic embedding model.
\newblock In {\em Advances in neural information processing systems}, pages
  2121--2129, 2013.

\bibitem{Gatys}
A.~Gatys, A.~Ecker, and M.~Bethge.
\newblock Image style transfer using convolutional neural networks.
\newblock In {\em Proceedings of the IEEE Conference on Computer Vision and
  Pattern Recognition}, pages 2414--2423, 2016.

\bibitem{Huang_2017_ICCV}
X.~Huang and S.~Belongie.
\newblock Arbitrary style transfer in real-time with adaptive instance
  normalization.
\newblock In {\em Proceedings of the IEEE International Conference on Computer
  Vision (ICCV)}, Oct 2017.

\bibitem{isola}
P.~Isola, J.~Zhu, T.~Zhou, and A.~Efros.
\newblock Image-to-image translation with conditional adversarial networks.
\newblock In {\em Proceedings of the IEEE Conference on Computer Vision and
  Pattern Recognition}, 2017.

\bibitem{jegou}
S.~J{\'e}gou, M.~Drozdzal, D.~Vazquez, A.~Romero, and Y.~Bengio.
\newblock The one hundred layers tiramisu: Fully convolutional densenets for
  semantic segmentation.
\newblock In {\em Proceedings of the IEEE Conference on Computer Vision and
  Pattern Recognition Workshops (CVPRW)}, pages 1175--1183. IEEE, 2017.

\bibitem{johnson}
J.~Johnson, A.~Alahi, and F.~Li.
\newblock Perceptual losses for real-time style transfer and super-resolution.
\newblock In {\em Proceedings of the European Conference on Computer Vision},
  pages 694--711. Springer, 2016.

\bibitem{li2017universal}
Y.~Li, C.~Fang, J.~Yang, Z.~Wang, X.~Lu, and M.-H. Yang.
\newblock Universal style transfer via feature transforms.
\newblock In {\em Advances in Neural Information Processing Systems}, pages
  385--395, 2017.

\bibitem{Li2017Demystifying}
Y.~Li, N.~Wang, J.~Liu, X.~Hou, Y.~Li, N.~Wang, J.~Liu, and X.~Hou.
\newblock Demystifying neural style transfer.
\newblock In {\em Twenty-Sixth International Joint Conference on Artificial
  Intelligence}, pages 2230--2236, 2017.

\bibitem{lian2016automatic}
Z.~Lian, B.~Zhao, and J.~Xiao.
\newblock Automatic generation of large-scale handwriting fonts via style
  learning.
\newblock In {\em SIGGRAPH ASIA 2016 Technical Briefs}, page~12. ACM, 2016.

\bibitem{lin2014microsoft}
T.-Y. Lin, M.~Maire, S.~Belongie, J.~Hays, P.~Perona, D.~Ramanan,
  P.~Doll{\'a}r, and C.~L. Zitnick.
\newblock Microsoft coco: Common objects in context.
\newblock In {\em European conference on computer vision}, pages 740--755.
  Springer, 2014.

\bibitem{liu2017}
M.~Y. Liu, T.~Breuel, and J.~Kautz.
\newblock Unsupervised image-to-image translation networks.
\newblock In {\em Advances in Neural Information Processing Systems}, pages
  700--708, 2017.

\bibitem{liu2016}
M.~Y. Liu and O.~Tuzel.
\newblock Coupled generative adversarial networks.
\newblock In {\em Advances in Neural Information Processing Systems 29}, pages
  469--477. 2016.

\bibitem{long}
J.~Long, E.~Shelhamer, and T.~Darrell.
\newblock Fully convolutional networks for semantic segmentation.
\newblock In {\em Proceedings of the IEEE Conference on Computer Vision and
  Pattern Recognition}, pages 3431--3440, 2015.

\bibitem{lyu}
P.~Lyu, X.~Bai, C.~Yao, Z.~Zhu, T.~Huang, and W.~Liu.
\newblock Auto-encoder guided gan for chinese calligraphy synthesis.
\newblock In {\em arXiv preprint arXiv:1706.08789}, 2017.

\bibitem{Mahendran2014}
A.~Mahendran and A.~Vedaldi.
\newblock Understanding deep image representations by inverting them.
\newblock pages 5188--5196, 2015.

\bibitem{mordvintsev}
A.~Mordvintsev, C.~Olah, and M.~Tyka.
\newblock Inceptionism: Going deeper into neural networks.
\newblock {\em Google Research Blog. Retrieved June}, 20(14), 2015.

\bibitem{Radford}
A.~Radford, L.~Metz, and S.~Chintala.
\newblock Unsupervised representation learning with deep convolutional
  generative adversarial networks.
\newblock In {\em Proceedings of the International Conference on Learning
  Representations}, 2016.

\bibitem{ronneberger}
O.~Ronneberger, P.~Fischer, and T.~Brox.
\newblock U-net: Convolutional networks for biomedical image segmentation.
\newblock In {\em Proceedings of the International Conference on Medical Image
  Computing and Computer-Assisted Intervention}, pages 234--241. Springer,
  2015.

\bibitem{Shrivastava}
A.~Shrivastava, T.~Pfister, O.~Tuzel, J.~Susskind, W.~Wang, and R.~Webb.
\newblock Learning from simulated and unsupervised images through adversarial
  training.
\newblock In {\em Proceedings of the IEEE Conference on Computer Vision and
  Pattern Recognition}, 2017.

\bibitem{Taigmand}
Y.~Taigman, A.~Polyak, and L.~Wolf.
\newblock Unsupervised cross-domain image generation.
\newblock In {\em arXiv preprint arXiv:1611.02200}, 2016.

\bibitem{Tenenbaum}
J.~Tenenbaum and W.~Freeman.
\newblock Separating style and content.
\newblock In {\em Proceedings of the Advances in neural information processing
  systems}, pages 662--668, 1997.

\bibitem{ulyanov}
D.~Ulyanov, V.~Lebedev, A.~Vedaldi, and V.~Lempitsky.
\newblock Texture networks: Feed-forward synthesis of textures and stylized
  images.
\newblock In {\em Proceedings of the International Conference on Machine
  Learning}, pages 1349--1357, 2016.

\bibitem{ulyanov2017improved}
D.~Ulyanov, A.~Vedaldi, and V.~Lempitsky.
\newblock Improved texture networks: maximizing quality and diversity in
  feed-forward stylization and texture synthesis.
\newblock In {\em Proc. CVPR}, 2017.

\bibitem{upchurch}
P.~Upchurch, N.~Snavely, and K.~Bala.
\newblock From a to z: supervised transfer of style and content using deep
  neural network generators.
\newblock In {\em arXiv preprint arXiv:1603.02003}, 2016.

\bibitem{wilmot}
P.~Wilmot, E.~Risser, and C.~Barnes.
\newblock Stable and controllable neural texture synthesis and style transfer
  using histogram losses.
\newblock {\em arXiv preprint arXiv:1701.08893}, 2017.

\bibitem{xian}
Y.~Xian, Z.~Akata, G.~Sharma, Q.~Nguyen, M.~Hein, and B.~Schiele.
\newblock Latent embeddings for zero-shot classification.
\newblock In {\em Proceedings of the IEEE Conference on Computer Vision and
  Pattern Recognition}, pages 69--77, 2016.

\bibitem{xu2009automatic}
S.~Xu, H.~Jiang, T.~Jin, F.~C. Lau, and Y.~Pan.
\newblock Automatic generation of chinese calligraphic writings with style
  imitation.
\newblock {\em IEEE Intelligent Systems}, 2009.

\bibitem{zhang2017multi}
H.~Zhang and K.~Dana.
\newblock Multi-style generative network for real-time transfer.
\newblock In {\em arXiv preprint arXiv:1703.06953}, 2017.

\bibitem{zhang2018drawing}
X.-Y. Zhang, F.~Yin, Y.-M. Zhang, C.-L. Liu, and Y.~Bengio.
\newblock Drawing and recognizing chinese characters with recurrent neural
  network.
\newblock {\em IEEE transactions on pattern analysis and machine intelligence},
  40(4):849--862, 2018.

\bibitem{zhang2018separating}
Y.~Zhang, Y.~Zhang, and W.~Cai.
\newblock Separating style and content for generalized style transfer.
\newblock In {\em Proceedings of the IEEE Conference on Computer Vision and
  Pattern Recognition}, 2018.

\bibitem{zhou2011easy}
B.~Zhou, W.~Wang, and Z.~Chen.
\newblock Easy generation of personal chinese handwritten fonts.
\newblock In {\em Multimedia and Expo (ICME), 2011 IEEE International
  Conference on}, pages 1--6. IEEE, 2011.

\bibitem{Zhu2017}
J.~Y. Zhu, T.~Park, P.~Isola, and A.~A. Efros.
\newblock Unpaired image-to-image translation using cycle-consistent
  adversarial networks.
\newblock In {\em Proceedings of the IEEE International Conference on Computer
  Vision (ICCV)}, Oct 2017.

\end{thebibliography}
}
\end{document}